\PassOptionsToPackage{dvipsnames}{xcolor}
\documentclass[10pt,twocolumn,letterpaper]{article}

\usepackage[pagenumbers]{cvpr} 

\definecolor{cvprblue}{rgb}{0.21,0.49,0.74}
\usepackage[pagebackref,breaklinks,colorlinks,allcolors=cvprblue]{hyperref}
\usepackage{amsmath}
\usepackage{algorithm}
\usepackage{algpseudocode}
\usepackage{colortbl}
\usepackage{array}  

\usepackage[accsupp]{axessibility}
\usepackage{tabularx}
\usepackage{multirow}
\usepackage{arydshln}
\usepackage{booktabs}
\usepackage{multirow}

\title{Fast Vision Mamba: Pooling Spatial Dimensions for Accelerated Processing}

\author{Saarthak Kapse$^{1, 2\thanks{Work completed during internship at Instiro}}$, Robin Betz$^1$, Srinivasan Sivanandan$^1$\\ \\
$^1$Insitro , $^2$Stony Brook University 
}

\begin{document}
\maketitle
\begin{abstract}
State Space Models (SSMs) with selective scan (Mamba) have been adapted into efficient vision models. Mamba, unlike Vision Transformers, achieves linear complexity for token interactions through a recurrent hidden state process. This sequential processing is enhanced by a parallel scan algorithm, which reduces the computational time of recurrent steps from $L$ sequential steps to $log(L)$ parallel steps with respect to the number of input tokens ($L$). In this work, we propose Fast Vision Mamba (FastVim), that further reduces the computational time of the SSM block by reducing the number of recurrent steps in Vision Mamba models while still retaining model performance. By alternately pooling tokens along image dimensions across Mamba blocks, we obtain a 2$\times$ reduction in the number of parallel steps in SSM block. Our model offers up to $72.5\%$ speedup in inference speed compared to baseline Vision Mamba models on high resolution (2048$\times$2048) images. Our experiments demonstrate state-of-the-art performance with dramatically improved throughput in a range of tasks such as image classification, cell perturbation prediction, segmentation, and object detection. Code is made available at \href{https://github.com/insitro/FastVim}{github.com/insitro/FastVim}
\end{abstract} 


\vspace{-0.5cm}

\section{Introduction}
\label{sec:introduction}

Recent developments in neural network architectures for computer vision tasks have used State Space Models~\cite{gu2021efficiently} (SSM) with selective scan (Mamba~\cite{mamba}) to enhance computational efficiency by replacing the quadratic complexity of self-attention in transformers~\cite{vaswani2017attention} with Mamba's linear complexity while retaining context dependence unlike recent SSMs~\cite{gu2021efficiently}. Models like Vision Mamba~\cite{vim} (Vim) and VMamba~\cite{vmamba} have shown that they can outperform their transformer-based counterparts, like Vision Transformer~\cite{vit} (ViT) and Swin~\cite{liu2021swin}, in vision tasks, while being particularly advantageous for tasks involving high-resolution images due to its efficient scaling. While Mamba supports content-based reasoning through selective scan in State Space Models (SSM), it cannot utilize efficient convolutions, necessitating a sequential recurrent approach that limits parallel processing. To address this, Mamba incorporates a parallel scan algorithm~\cite{blelloch1990prefix}, reducing sequential steps to a \textit{lower bound} of logarithmic scale~\cite{smith2022simplified} with respect to the number of tokens. While this approach significantly reduces sequential steps, in the vision domain, the number of tokens scale \textit{quadratically with increasing resolution}. Consequently, this results in a quadratic increase in the number of sequential recurrent steps translating to 2× increase in the number of parallel steps when using parallel scan, which challenges throughput in high-resolution imaging.

\begin{figure}[!t]
\centering
    \includegraphics[width=1\linewidth]{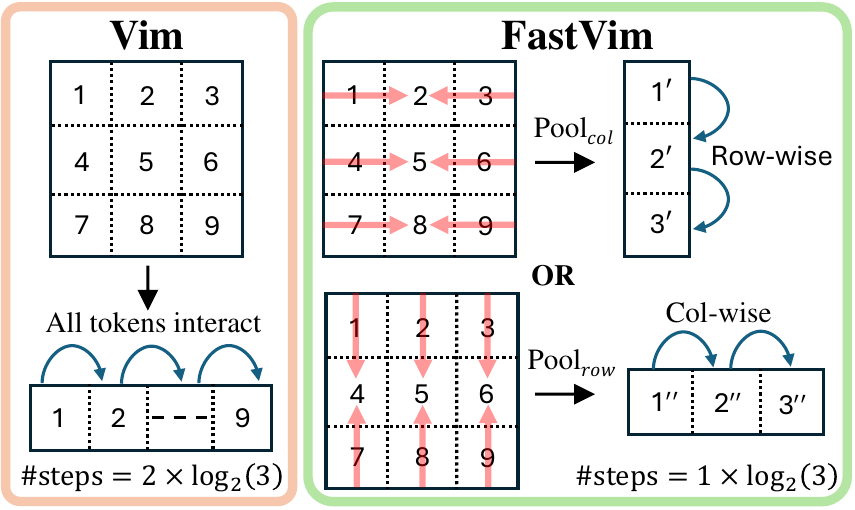} 
    \caption{
    FastVim accelerates Vim by mean pooling tokens across columns or rows, transforming token scaling from quadratic to linear with resolution. FastVim requires $log(h)$ parallel steps, compared to Vim's $log(h^2)$ parallel steps in Mamba's contextualization module SSM where $h$ is the number of tokens along height or width of the image with $L = h^2$ token inputs to the model.
    }
     \vspace{-6pt}
    \label{fig:teaser}
\end{figure}

In this work, we explore the possibility of scaling the number of recurrent computations in Vision Mamba linearly with image resolution, as opposed to scaling quadratically. We do this by applying average pooling across one dimension of the 2D token grid before the recurrent SSM block. This raises the question: \textit{Can we reduce the number of recurrent computations in Vision Mamba without compromising model performance?}

The answer is yes. In this paper, we utilize a simple, parameter-free technique of average pooling to reduce the number of recurrent steps in Vision Mamba while maintaining strong predictive power. 
An important consideration is alternating the pooling dimensions (as shown in Fig.~\ref{fig:teaser}) across stacked Mamba blocks. Pooling tokens across column (Pool$_{col}$) prevents interaction of tokens in a row and similarly across row pooling (Pool$_{row}$) prevents interaction of tokens in a column; thus, alternation ensures all tokens interact implicitly across multiple blocks. We empirically demonstrate that this alternation is a \textit{necessity} for achieving high performance in visual encoding, not just a desirable feature.

Our method, FastVim, is a purely Mamba-based neural network architecture (built on Vim) that uses pooling to accelerate contextualization in SSM scans. As shown in Fig.~\ref{fig:model}, in each forward and backward scan branch, mean pooling is applied after a 1D convolution layer to compress tokens across rows or columns, resulting in a one-dimensional token grid. These compressed tokens are projected to input-dependent parameters via a linear layer for selective scan, followed by interaction in the SSM module. The output is then repeated to restore the original token grid before the skip connection, followed by norm layer. Thus, across blocks, the number of tokens remains unchanged: they are compressed with pooling before the SSM scan and decompressed with repetition afterward, as shown in Fig~\ref{fig:model}. To alternate pooling dimensions and align 1D convolution direction with the SSM scanpath direction, we transpose the token grids every block. We further investigated whether our pooling approach is effective in ViT, and experimental results demonstrate that while it works in Vim, it fails in ViT, highlighting the need to further study Mamba's contextualizing capabilities versus Transformers. 

To extend our approach across other domains, we propose the following two adaptations of FastVim: FastMaskVim - incorporating masking in FastVim for applications like Mask Autoencoders~\cite{mae} (MAE), DINOv2~\cite{dinov2}, and pathology datasets~\cite{chen2024towards} having non-regular grids, and FastChannelVim - utilizing per-channel tokenization as introduced by ChannelViT~\cite{channelvit}, beneficial for datasets like microscopy cell~\cite{chandrasekaran2023jump} and satellite imaging. Remarkably, we trained the most effective pure Mamba-based monolithic visual encoder using our pooling method with MAE on ImageNet-1K~\cite{imagenet} to date, achieving state-of-the-art performance (SOTA). In per-channel tokenization task, FastChannelVim showed phenomenal gains in accuracy over ChannelViT baselines on microscopy image classification task, demonstrating the benefits of our method for long token sequences in vision. In summary, our main contributions are:

\begin{itemize}
    \item FastVim, a Vim-based architecture that utilizes average pooling, achieving 1× parallel steps in logarithmic scale from 2×, translating to $72.5\%$ speedup in overall framework.

    \item FastVim is adapted into FastMaskVim and FastChannelVim, extending its utility to applications with irregular grids and multi-channel imaging, respectively.

    \item Our methods set a new SOTA of \textit{86.7}\% on ImageNet-1k~\cite{imagenet} (with MAE pretraining) with a Mamba-based encoder and show substantial improvements over transformer baselines in long token sequence modeling in per-channel tokenization on microscopy imaging by $8.3\%$.
    
\end{itemize}

\vspace{-0.1cm}
\section{Preliminaries}
\label{sec:preliminaries}

\begin{figure*}[t]
\centering
    \includegraphics[width=1\linewidth]{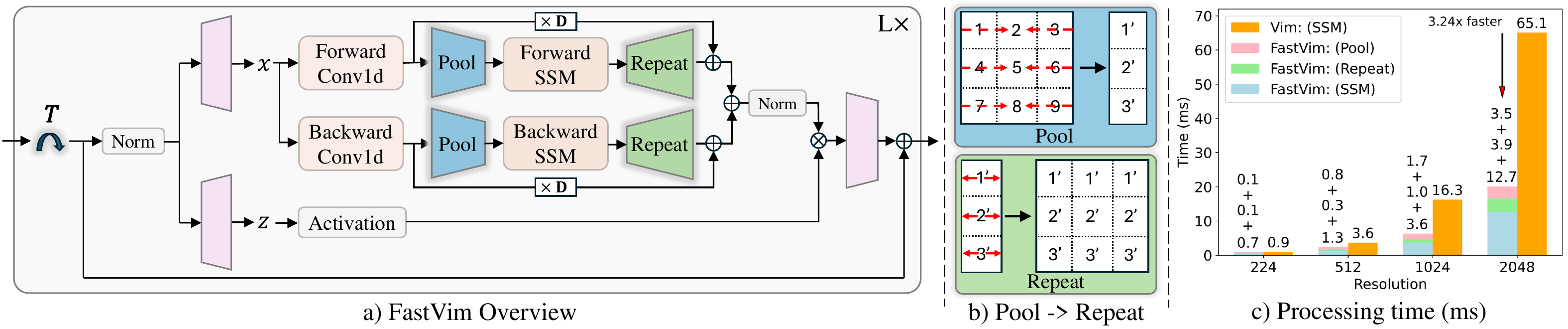} 
    \caption{\textbf{Overview of FastVim}: Input image tokens are fed to norm and expansion layers, then output $x$ is transposed ($T$) every block for alternate pooling of rows and columns. Tokens are pooled post-\texttt{Conv1D}, processed by SSM, and decompressed before skip-connection ($\mathbf{D}$ in eq.~\ref{eq:dtm}). Note that the flattened tokens are reshaped into a 2D grid prior to the transpose and pooling layers, and are flattened again after these operations. In c) we illustrate the comparison of Forward SSM + Backward SSM inference time in one layer of Vision Mamba with Forward SSM + Backward SSM + pooling + repeat inference time in one layer of Fast Vision Mamba. We observe that with increase in resolution, FastVim needs significantly less time than Vim for contextualization module (further detailed in Supplement~\ref{additional_throughput}).}
    \vspace{- 4pt}
    \label{fig:model}
\end{figure*}

\textbf{State Space Models (SSMs)} are mathematical frameworks that model continuous-time sequences by transforming an input sequence $x(t) \in \mathbb{R}$ into an output sequence $y(t) \in \mathbb{R}$ via a hidden state $h(t) \in \mathbb{R}^N$, as governed by:

\begin{equation}
\begin{aligned}
h^{'}(t) &= \mathbf{A} h(t) + \mathbf{B} x(t), \\
y(t) &= \mathbf{C} h(t) + \mathbf{D} x_t,
\end{aligned}
\label{eq:ctm}
\end{equation}

where $\mathbf{A} \in \mathbb{R}^{N \times N}$ describes how the current state evolves, $\mathbf{B} \in \mathbb{R}^{N \times 1}$ describes how the input influences the state, $\mathbf{C} \in \mathbb{R}^{1 \times N}$ describes how the current state translates to the output, $\mathbf{D} \in \mathbb{R}$ describes how the input directly influences the output, acting as a skip connection, and $N$ being number of states.

To be applied on discrete sequence datasets, SSMs are discretized using zero-order hold over a sampling interval $\Delta$, resulting in discrete parameters $\overline{\mathbf{A}}$ and $\overline{\mathbf{B}}$:

\begin{equation}
\begin{aligned}
\overline{\mathbf{A}} &= e^{\Delta \mathbf{A}}, \quad 
\overline{\mathbf{B}} &= (\Delta \mathbf{A})^{-1} \left( e^{\Delta \mathbf{A}} - \mathbf{I} \right) \Delta \mathbf{B}
\end{aligned}
\label{eq:discrete}
\end{equation}

The discrete-time SSM equations are then modified as:

\begin{equation}
\begin{aligned}
h_t &= \overline{\mathbf{A}} h_{t-1} + \overline{\mathbf{B}} x_t, \quad y_t &= \mathbf{C} h_t + \mathbf{D} x_t
\end{aligned}
\label{eq:dtm}
\end{equation}

The above equation can be computed like a recurrent neural network, viewing $h_t$ as a hidden state with transition matrix $\overline{\mathbf{A}}$. However, since it is not practical for training on modern hardware due to sequential processing, the well-known connections between linear time-invariant (LTI) SSMs (eq.~\ref{eq:dtm}) and continuous convolutions can be used. Specifically, the eq.~\ref{eq:dtm} can be computed as $y = \overline{\mathbf{K}} * x$, where $\overline{\mathbf{K}}$ is the SSM convolution kernel. However, computing $\overline{\mathbf{K}}$ is computationally prohibitive due to repeated matrix multiplication of $\overline{\mathbf{A}}$ (for more depth, please see Linear State Space Layer (LSSL)~\cite{gu2021combining}). To address this, the Structured State Space sequence model (S4)~\cite{gu2021efficiently} proposed a novel parameterization of $\overline{\mathbf{A}}$, making it significantly faster and less memory consuming, while exceeding the LSSL’s performance empirically.

\noindent \textbf{Selective State Space Models} termed as Mamba~\cite{mamba} enhance S4 further by allowing input-dependent parameters, enabling the model to capture richer contextual information. Mamba modifies the parameters $\mathbf{B}$, $\mathbf{C}$, and $\Delta$ as functions of the token sequence $x \in \mathbb{R}^{B \times L \times D}$, where $B$ is the batch size, $L$ is the sequence length, and $D$ is the token embedding dimension, as follows:

\begin{equation}
\begin{aligned}
\mathbf{B} &= \text{s}_{B}(x), \quad \mathbf{C} = \text{s}_{C}(x), \\
\mathbf{\Delta} &= \tau_\Delta(\text{Parameter} + \text{s}_{\Delta}(x)),
\end{aligned}
\end{equation}

where $\text{s}_{B}(x) = \text{Linear}_{N}(x)$, $\text{s}_{C}(x) = \text{Linear}_{N}(x)$, $\text{s}_{\Delta}(x) = \text{Broadcast}_{D}(\text{Linear}_1(x))$, $\tau_\Delta = \text{softplus}$, and $N$ being number of states, while $\text{Linear}_{d}$ is a parameterized projection to dimension $d$. This allows the model to adaptively modulate state transitions since the parameters are now token-dependent, thus improving its ability to model complex sequences. 

However, this comes at the cost of not being able to use the convolution operation like in S4 since the parameters are no longer linear time-invariant. This results in two challenges: the sequential nature of recurrence and the large memory usage. To address the latter, Mamba proposed a hardware-aware implementation that leads to significant speedup. For sequential recurrence, Mamba accelerated it by employing parallel scanning~\cite{smith2022simplified}, thus reducing the steps from $L$ sequential steps to a logarithmic scale, $\log(L)$ parallel steps.

\textbf{Vim} extends Mamba, which was originally designed for 1-D sequences, to handle visual data by transforming 2-D imaging datasets in line with adaptation similar to transformers in Vision Transformers. An input image of size $(H \times W \times C)$ is divided into flattened patches of size $(P \times P \times C)$, where $H$ is the height, $W$ is the width, $C$ is the number of channels, and $P$ is the patch size. This results in $L$ patches, with $L = \frac{H \times W}{P^2}$. These patches are projected into tokens of dimension $D$ using a linear layer, forming a token sequence $x \in \mathbb{R}^{L \times D}$.
Vim further proposed using two SSM modules in each layer, namely Forward SSM and Backward SSM, along with a 1-D causal convolution before both SSMs to allow for bi-directional contextualization needed for understanding non-causal imaging data. For more details about the overall architecture we refer the readers to~\cite{vim} and Fig.~\ref{fig:model} ignoring our modifications of transpose $T$, pool, repeat, and post-SSM norm operations.

\vspace{-0.1cm}

\section{Method}
\label{sec:method}

In this section, we present the details of our Fast Vision Mamba (FastVim), providing an overview in Fig.~\ref{fig:model}. A detailed description of our proposed pooling method, designed to accelerate contextualization in Vision Mamba, is provided in Sec.~\ref{subsec:pooling}. In Sec.~\ref{subsec:masking}, we explore extensions to masking paradigms, whereas in Sec.~\ref{subsec:channelmodeling} we illustrate the extension to the domain of per-channel tokenization modeling~\cite{channelvit}.

\subsection{Spatial Pooling for faster contextualization}
\label{subsec:pooling}

In this paper, we propose a novel method to reduce the number of recurrent steps in Vim through spatial pooling (FastVim). Specifically, as detailed in Algorithm 1, we propose mean pooling across one spatial dimension of a 2-D image's token grid ($x$). Suppose we have a square grid where $h=w$ and $L=h \times w=h^2$ (where $h=H/P$ and $w=W/P$ are spatial dimensions of token sequence $x$ before flattening), this pooling reduces the sequence length to $h$ from $h^2$, resulting in a $1\times$ parallel steps (when using parallel scan) in FastVim ($\log(h)$) compared to $2\times$ parallel steps in Vim ($\log(h^2)$). Note that we use a square grid for a simpler example, but FastVim is generalizable to any image dimensions. This approach fits within the sparse contextualization paradigm because, instead of all tokens interacting, only pooled tokens interact with each other across one spatial dimension. Following the scan operation, the output is repeated to get back the sequence of size $h^2$. 
Average pooling is used as a default; variants with max and attention pooling are in the Supplement.

\begin{algorithm}
\caption{SSM + Selection + Spatial Pooling}
\textbf{Input:} $x : (B,L,D)$,  where $B = \text{batch size}$, $L = h \times w$, and $D = \text{embedding dimension}$ \\
\textbf{Output:} $y : (B,L,D)$
\begin{algorithmic}[1]
    \State $\textbf{A} : (D,N) \gets$ Parameter
    \Statex \Comment{Represents structured $N \times N$ matrix, where $N = \text{number of states}$}
    \State $x : (B,h,w,D) \gets \text{reshape}(x)$
    \State $x_{\text{pooled}} : (B,h,1,D) \gets  \text{pool}(\textbf{$x$}[:, :, :w, :])$ 
    \Statex \Comment{Pool spatial dimension}

    \State $\boldsymbol{B} : (B,h,1,N) \gets s_B(x_{\text{pooled}})$
    \State $\boldsymbol{C} : (B,h,1,N) \gets s_C(x_{\text{pooled}})$
    \State $\boldsymbol{\Delta} : (B,h,1,D) \gets \tau_\Delta(\text{Parameter} + s_\Delta(x_{\text{pooled}}))$

    \State $\bar{\textbf{A}}, \bar{\textbf{B}} : (B,h,1,D,N) \gets$ discretize($\Delta, A, B$)
    
    \State $y_{\text{pooled}} : (B,h,1,D) \gets \text{SSM}(\bar{\textbf{A}}, \bar{\textbf{B}}, \textbf{C})(x_{\text{pooled}})$
    \Statex \Comment{Time-varying: recurrence (scan) only}

    \State $y : (B,h,w,D) \gets \text{repeat}(y_{\text{pooled}}, \text{along } w)$

    \State $y : (B,L,D) \gets \text{reshape}(y)$
    
    \State \textbf{return} $y$
\end{algorithmic}
\end{algorithm}

Intuitively, pooling might lead to inadequate contextualization of tokens in a row when pooling tokens across columns (Pool$_{col}$), and similarly for tokens in a column when pooling tokens across rows (Pool$_{row}$). We address this issue by alternating the pooling operation across rows and columns across layers in FastVim. This enhances effective interactions among pooled tokens in different rows \textit{(row-wise interaction)} in Pool$_{col}$ and pooled tokens in different columns \textit{(col-wise interaction)} in Pool$_{row}$ (as shown in Fig.~\ref{fig:teaser}). We empirically demonstrate that this adjustment is crucial for achieving performance comparable to the baseline Vim. As shown in Fig.~\ref{fig:model}, this is carried out using a transpose of the token grid at every block, as we want to apply a 1D-conv in same direction as SSM scan. In practice as we use transpose, we always pool tokens across columns.

\subsection{FastMaskVim: Incorporating Masking}
\label{subsec:masking}

So far, we have described FastVim in the context of a regular token grid of size $h \times w$. However, this approach cannot be directly utilized when faced with an irregular grid, a situation often encountered in scenarios involving masked tokens such as in Masked Autoencoders~\cite{mae, zhouhypermae} (MAE) and DINOv2~\cite{dinov2}, or in multiple instance learning~\cite{abmil, simil} (MIL) in pathology, where tissue samples can contain gaps. To enable FastVim to function effectively in such domains, we need to modify the pooling and transpose operations.

\begin{figure}[!h]
\centering
    \includegraphics[width=0.7\linewidth]{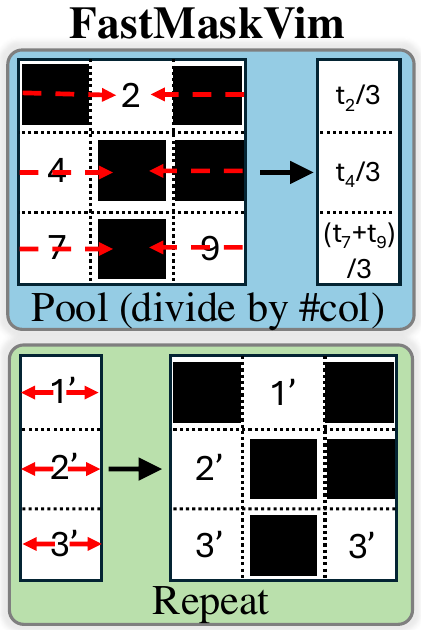} 
    \caption{Illustration of pooling and repeat operations in FastMaskVim. Instead of naive mean pooling of tokens in a row, we add the tokens and then divide it by number of columns in the token grid. Similarly when alternatively pooling tokens in a column.}
\label{fig:fastmaskvim_teaser}
\end{figure}

Specifically, instead of using a simple transpose operation on the token grid, we employ advanced indexing techniques to transpose a sparse token grid of shape $h \times w$, but only including the unmasked tokens. For pooling, we sum the tokens in each row and then divide by the number of columns, i.e., $w$, instead of naively performing mean pooling (see Fig.~\ref{fig:fastmaskvim_teaser}), as mean pooling could result in the loss of information regarding the number of tokens present in the row. These simple modifications have proven effective, as is  demonstrated by the MAE-pretrained FastMaskVim in Sec.~\ref{subexp:mae}.

\subsection{FastChannelVim: Per-Channel tokenization}
\label{subsec:channelmodeling}

In 2-D imaging datasets, a region of size $P \times P \times C$ is typically projected into a single token of dimension $D$, where $P$ is the patch size and $C$ is the number of channels, thus forming a token sequence $x \in \mathbb{R}^{L \times D}$ for $L$ tokens. However, this tokenization approach has been shown to be inadequate for certain imaging modalities where per-channel information is highly complementary, such as in microscopy cell imaging and satellite imaging, unlike the RGB channels in natural images. As established by ChannelViT~\cite{channelvit}, per-channel tokenization can address this limitation, though at the cost of increasing the number of tokens by a factor of $C$, thus forming a token sequence $x \in \mathbb{R}^{(L.C) \times D}$. In this paradigm, channel embedding is added along with position embedding to preserve order information.

\begin{figure}[!h]
\centering
    \includegraphics[width=1\linewidth]{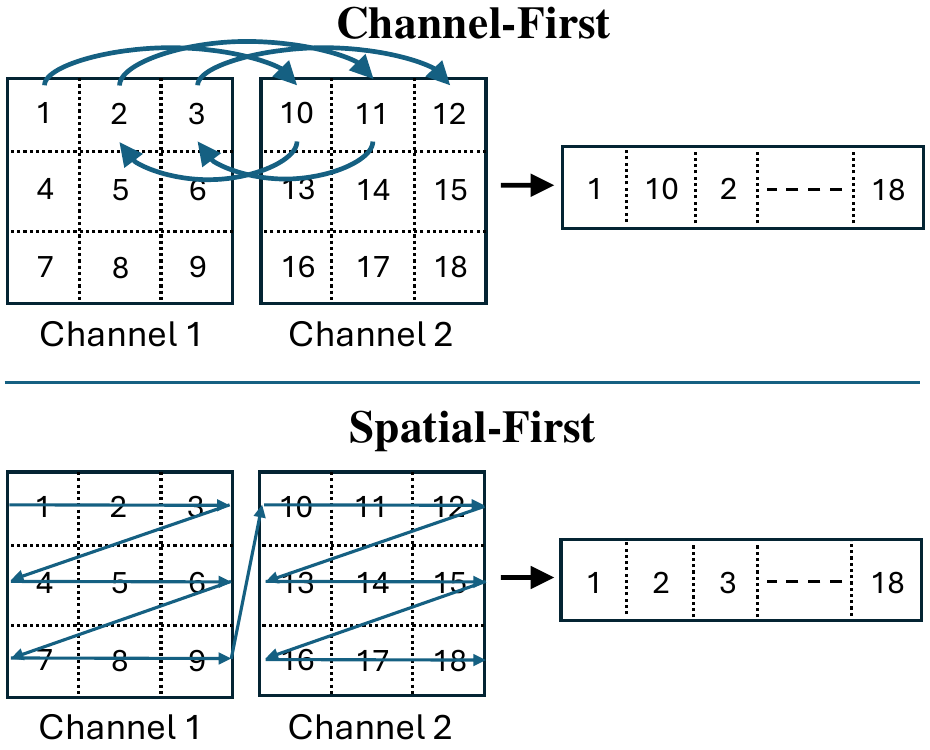} 
    \caption{Illustration of flattened scanpath options available following per-channel tokenization in ChannelVim due to sequential nature of Mamba.}
\label{fig:channelvim_scanpath}
\end{figure}

Building on the benefits (performance and efficiency) of Mamba over Transformers in long sequence settings, we introduce an extension of Vim with per-channel tokenization, which we term ChannelVim. To implement this extension, we must address two key considerations due to the sequential nature of SSM scan in Mamba, in contrast to the set-like, permutation-invariant nature of self-attention in transformers. First, for the scan path, as illustrated in Fig.~\ref{fig:channelvim_scanpath}, we have two options: we can either traverse across all spatial tokens within a channel and then proceed to the next channel (spatial-first approach), or we can traverse across all channels at a given spatial position and then move to the next spatially adjacent position and repeat (channel-first approach). Second, it has been shown that hierarchical channel sampling (HCS), where some channels are randomly dropped during training, improves performance~\cite{channelvit}. We incorporate such HCS in ChannelVim. However unlike the original implementation, the output of the HCS module needs to be sorted, as order of channels matters in sequential modeling. We provide thorough evaluation of the effect of both above mentioned considerations in the Supplement.~\ref{additional_jumpcp_implementation_section}.

\begin{figure}[!h]
\centering
    \includegraphics[width=0.7\linewidth]{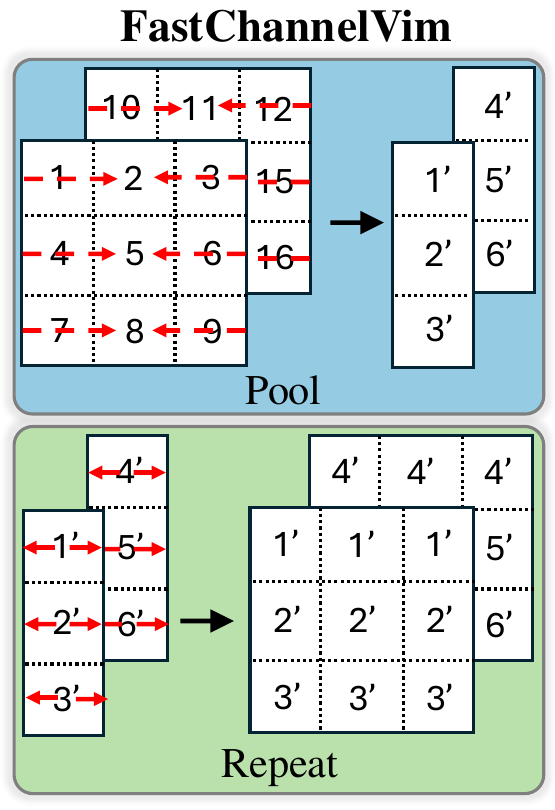} 
    \caption{Illustration of pooling and repeat operations in FastChannelVim. These two operations are performed independently for each channel in per-channel tokenization paradigm.}
\label{fig:fastchannelvim_teaser}
\end{figure}

\begin{figure}[!h]
\centering
    \includegraphics[width=1\linewidth]{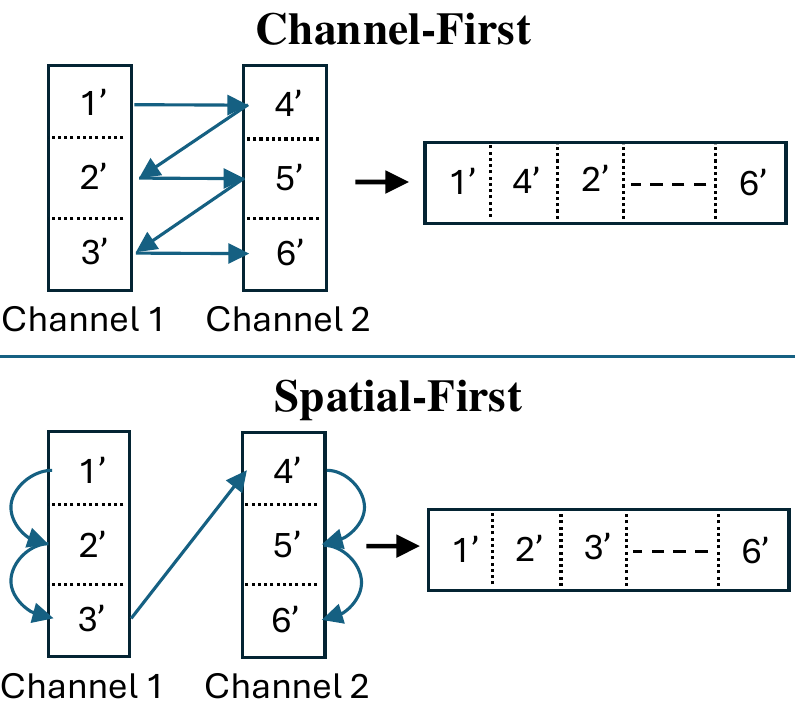} 
    \caption{Illustration of flattened scanpath options available following per-channel tokenization in FastChannelVim due to sequential nature of Mamba.}
\label{fig:fastchannelvim_scanpath}
\end{figure}

Finally, we adapt our FastVim to this domain, which we term FastChannelVim. In the main paper, we explore compressing tokens only across the spatial dimensions  (see Fig.~\ref{fig:fastchannelvim_teaser}). Thus, for each scan operation, we input either $h \times C$ (see Fig.~\ref{fig:fastchannelvim_scanpath}) or $w \times C$ tokens, instead of the entire $h \times w \times C$ tokens. In Supplement~\ref{additional_jumpcp_implementation_section}, we also explore compressing across the channel dimension.

\vspace{-0.1cm}

\section{Experiments and Results}
\label{sec:experiments}

\subsection{Image Classification}

\noindent \textbf{Settings.}
We conduct training on the ImageNet-1k dataset~\cite{imagenet} consisting of 1.28M training images, and utilize the 50K ImageNet-1k validation images for evaluation. We follow exact training settings from~\cite{vim}, i.e. we train our models for 300 epochs using a batch size of 1,024, the AdamW optimizer, and EMA. A cosine annealing learning rate schedule with an initial value of $1\times10^{-3}$, a 5-epoch warmup period, and a weight decay of 0.05 is used. For data augmentation, we apply standard techniques such as random cropping, horizontal flipping, label-smoothing regularization, mixup, and random erasing. For FastVim-B we use a higher drop path rate of 0.4 instead of default 0.05 to avoid over-fitting.

\definecolor{lightblue}{RGB}{173, 216, 255}
\begin{table}[ht]
    \caption{Classification benchmarks on \textbf{ImageNet-1k}~\cite{imagenet} dataset. All models are trained from scratch on Image size of $224\times224$. $\dag$ denotes we extend the training of Vim to base-size model. T refers to Tiny, S to Small, and B to Base size models.} 
\vspace{-13pt}
    \begin{center}
    \resizebox{0.80\columnwidth}{!}{
    \begin{tabular}{lccc}
        \toprule
        Model & \#Params &  FLOPs & Top-1 \\

         & (M) & (G) & (\%) \\
        
        \midrule

        Vim-T~\cite{vim} & 7M & 1.8G & 76.1 \\
        Vim-S~\cite{vim} & 26M & 5.9G & 80.5 \\

        Vim-B~\cite{vim}$\dag$ & 98M & 20.9G & 80.7 \\
        Vim-B w/ LN$\dag$ & 98M & 21.0G & 82.6 \\

\midrule
        \rowcolor[HTML]{E4E8FF}  
        FastVim-T & 7M & 1.17 & 75.4 \\
        
        \rowcolor[HTML]{E4E8FF}  
        FastVim-S & 26M & 4.43 & 81.1 \\
        
        \rowcolor[HTML]{E4E8FF}  
        FastVim-B & 98M & 17.23 & 82.6 \\

        \bottomrule
    \end{tabular}
}
\end{center}
\vspace{-8pt}
    \label{tab:imagenet_results}
\end{table}

\noindent \textbf{Results.} As seen in Table~\ref{tab:imagenet_results}, our proposed FastVim models perform on-par to the baseline Vim models across all model sizes. Since our approach is parameter-free, achieving comparable performance demonstrates that the proposed token pooling of FastVim still maintains sufficient token interaction across SSM blocks in Vim. Our empirical findings indicate that Vim can be effectively trained using a much sparser interaction of token. We report performance comparison with ViTs and other Mamba baselines in Supplement Table.~\ref{tab:additional_main_results}. It's important to mention that our research complements recent developments in the Vision Mamba field, such as VMamba~\cite{vmamba}, MambaVision~\cite{mambavision}, and GroupMamba~\cite{groupmamba}. Our pooling-based token reduction technique can be easily integrated within these methods to further enhance the speed of those high-performing architectures.

\begin{figure}[t]
\centering
    \includegraphics[width=0.8\linewidth]{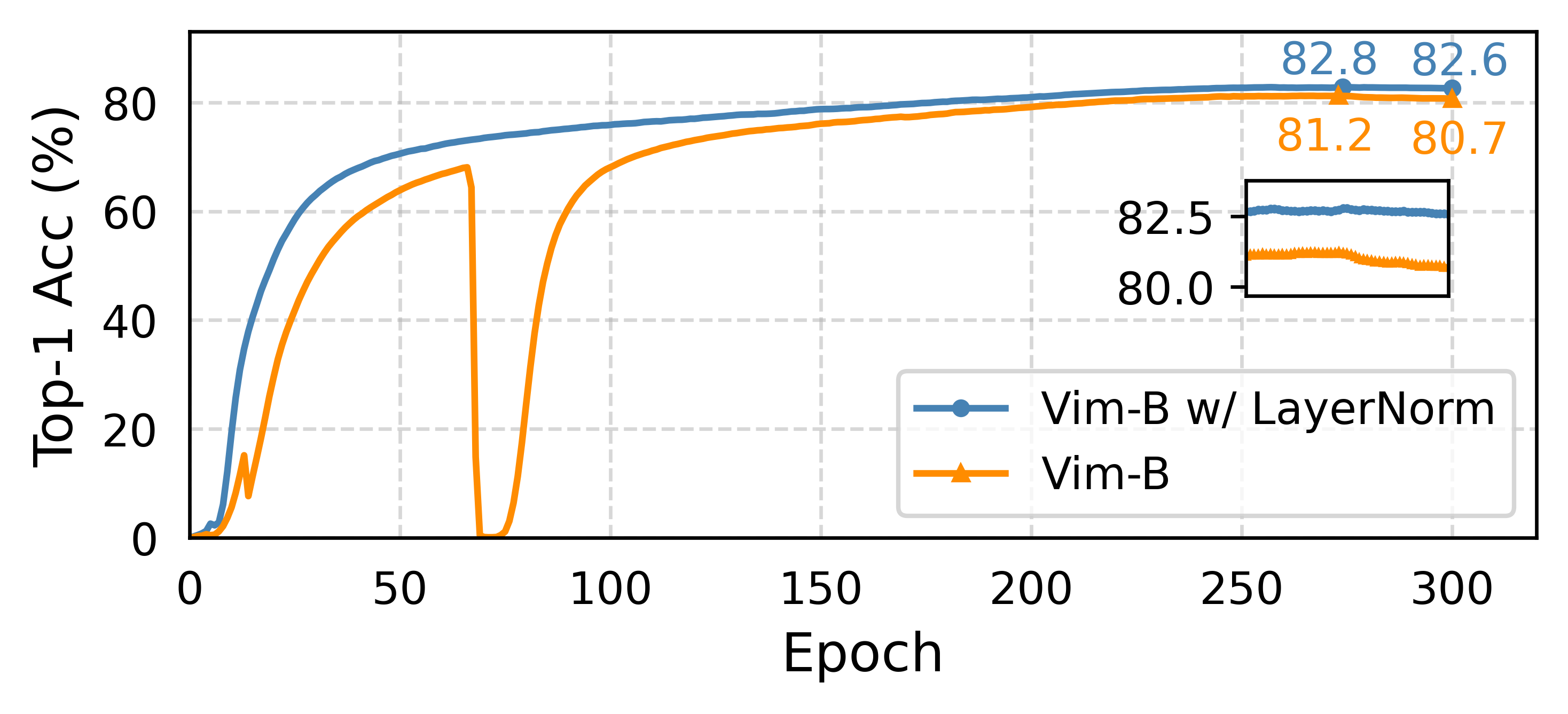}
    \caption{Stability Issue in Vim-B on ImageNet-1k.
    }
     \vspace{-10pt}
    \label{fig:stability}
\end{figure}

\noindent \textbf{Stability Issue in Vim.} As shown in Fig.~\ref{fig:stability}, we trained Vim-B from scratch on ImageNet-1k using default settings and a drop path rate of 0.4. During training, we encountered the issue of loss spikes, which caused instability. At the convergence, the model reached a peak accuracy of 81.2\%, consistent with recent work~\cite{ren2024autoregressive}. Inspired by the findings in~\cite{vmamba, jamba}, we experimented with adding LayerNorm post SSM scan. This modification resulted in much more stable training, overcoming the instability and ultimately improving the performance of Vim-B to 82.6\%. Our FastVim-B, which already incorporates LayerNorm post SSM operation (see fig.~\ref{fig:model}), achieved the same performance of 82.6\%. This illustrates the necessity of including the extra norm in the Vim and FastVim modules, aligning with the presence of two normalization layers in each transformer block, as well as in VMamba~\cite{vmamba}. In our experience, this spikes starts emerging from the small size Vim.

\subsection{Efficiency Analysis}
\label{subexp:throughput}

\begin{figure}[!h]
\centering
    \includegraphics[width=0.9\linewidth]{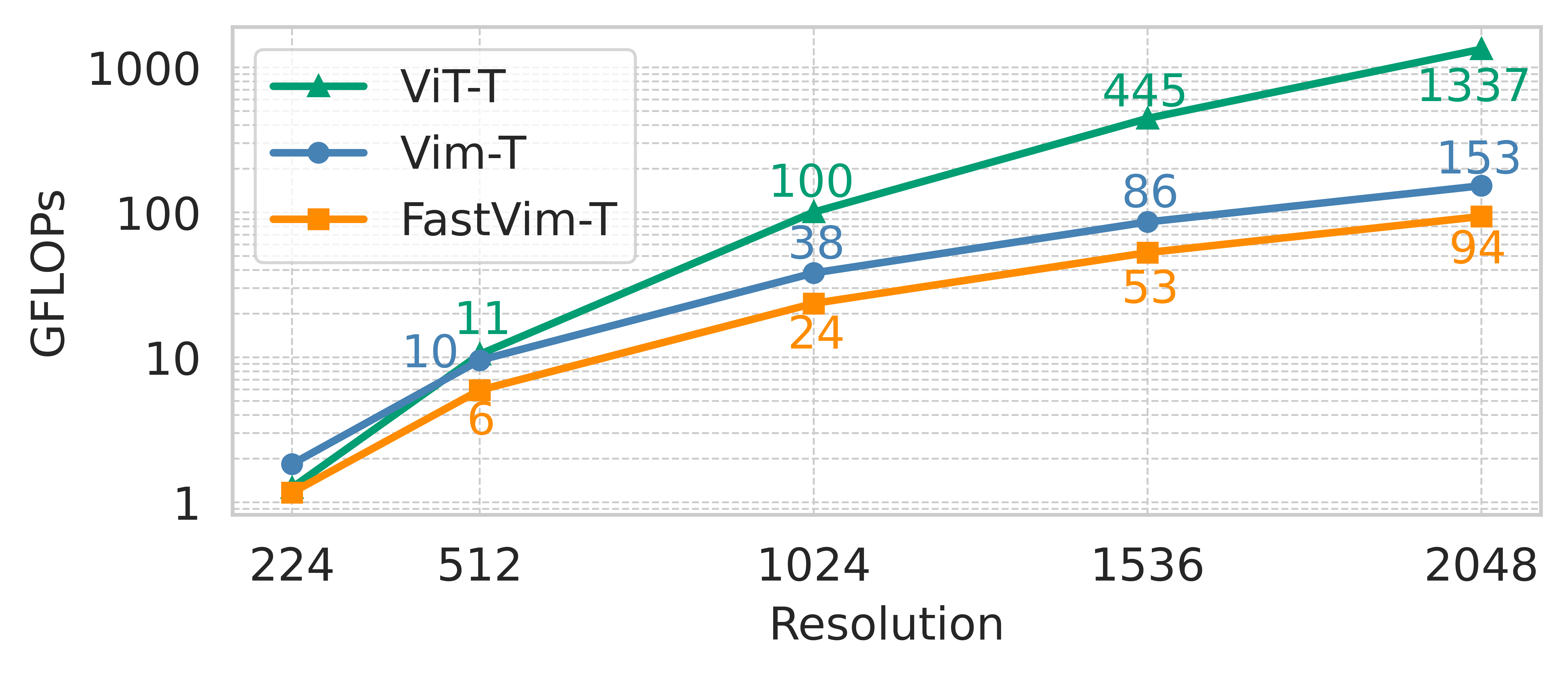} 
    \vspace{-8pt}
    \caption{
    Comparison of FLOPs (G) for FastVim, Vim, and ViT across different resolutions. 
    }
    \vspace{-4pt}
    \label{fig:flops}
\end{figure}

Here, we demonstrate the reduction in FLOPs and the increase in throughput achieved by our FastVim compared to Vim. In Fig.~\ref{fig:flops}, we compare the FLOPs requirements of FastVim, Vim, and ViT. At a lower resolution of 224, Vim demands the most operations, whereas ViT and FastVim have similar computational needs. As the resolution increases, ViT's computational requirements grow quadratically, while both Vim and FastVim scale linearly, with FastVim using up to 38\% fewer FLOPs. Notably, within a Mamba block, all components scale linearly in terms of FLOPs with the number of tokens, leading to a quadratic increase with respect to resolution for vision tasks. FastVim optimizes computations exclusively in the SSM, reducing its scaling to linear with respect to resolution. As a result, the other layers remain unchanged and maintain the same quadratic scaling as in Vim. Consequently, the overall FLOPs reduction in FastVim compared to Vim does not widen significantly with increasing resolution. The computational savings become more apparent at the SSM level, but this widening effect is muted at the block level, with FastVim-T using 35\% fewer FLOPs at 224 resolution and 38.5\% fewer FLOPs at 2048 resolution compared to Vim-T.

\begin{figure}[!h]
\centering
    \includegraphics[width=1\linewidth]{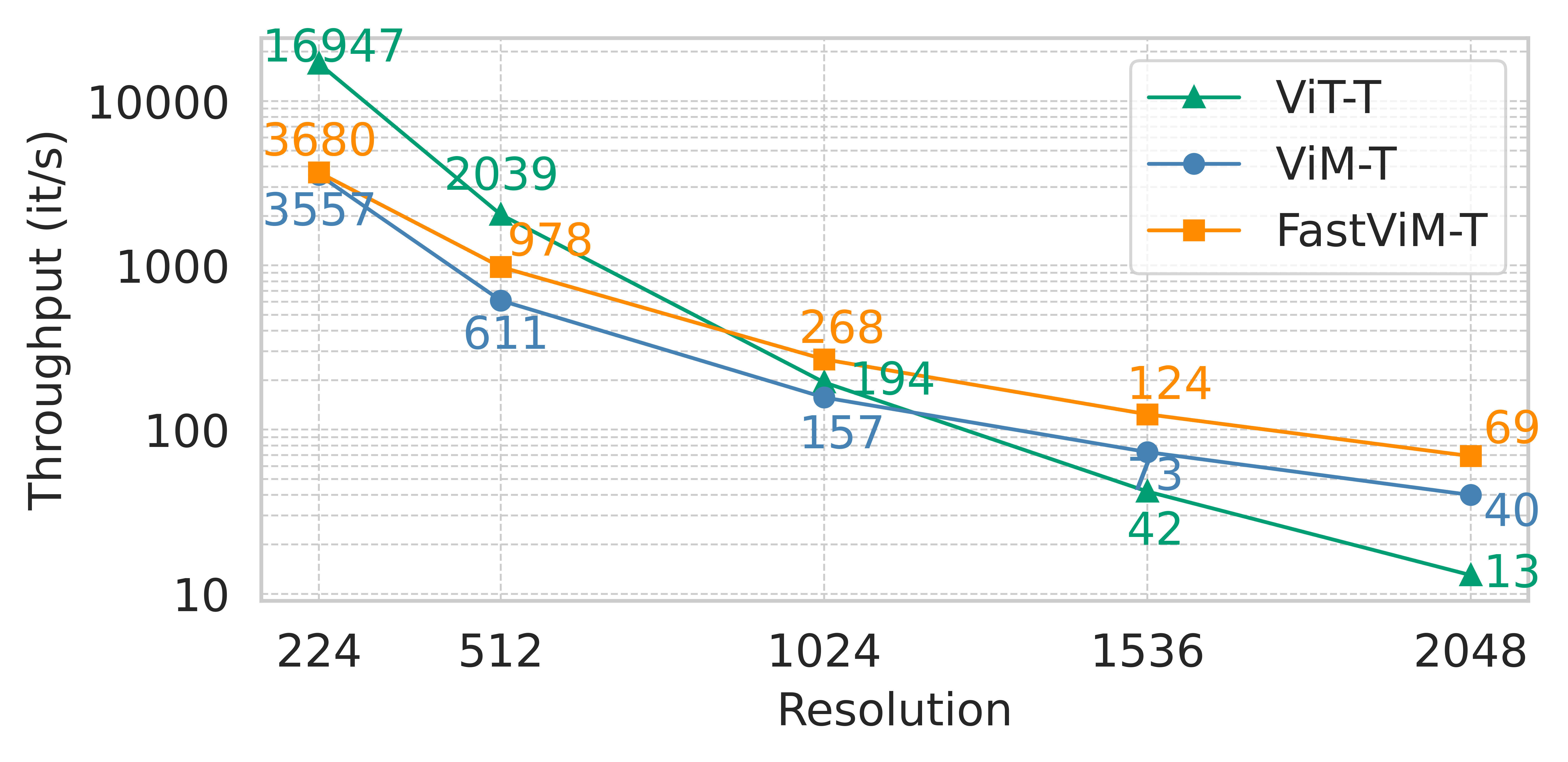} 
    \vspace{-12pt}
    \caption{
    Comparison of Inference Throughput (it/s) for FastVim, Vim, and ViT across different resolutions. Tested on H100 GPUs with batch size of 128.
    }
    \vspace{-6pt}
    \label{fig:throughput}
\end{figure}

However, as observed in the case of throughput (Fig.~\ref{fig:throughput}), the gap between Vim and FastVim widens with increasing resolution, as FastVim's throughput relative to Vim's continually improves. Based on our observations (detailed in Supplement Table~\ref{tab:dissecting_SSM_time}), we found that the SSM scan time remains nearly constant for FastVim across resolutions from 224 to 2048, whereas it increases by up to 74$\times$ with an 8$\times$ increase in resolution (64$\times$ increase in tokens) in Vim. The reasoning behind this observation is that even at an image size of 2048, after tokenization and pooling, FastVim's SSM scan processes only 128 tokens, compared to 196 tokens in Vim's SSM scan at a much lower resolution of 224. Unlike no contribution of pooling and repeating operations to FLOPs, these operations incur overhead processing time in FastVim. However, with increasing resolution, the rapid decrease in SSM scan processing time in FastVim overpowers the increasing overhead (see Supplement Table~\ref{tab:dissecting_SSM_time}). Thus, at a resolution of 2048, the time taken by the Forward and Backward SSM layer in a block shows a 324\% speedup in FastVim compared to Vim, translating to nearly a 72.5\% speedup in the overall model, as the other MLP and gating layers remain unchanged in both Vim and FastVim. We also observe that, at a resolution of 1024 and beyond, our method outperforms ViT in terms of speed, while requiring less than four times the FLOPs at 1024. This translates to a significantly lower memory requirement, with the computation gap widening rapidly at higher resolutions (see Fig.~\ref{fig:flops}). Thus, at high resolution, our solution is not only faster than both Vim and ViT, but it also consumes substantially less memory than ViT. Note that we used LayerNorm post-SSM for both Vim and FastVim (see Fig.~\ref{fig:stability}), and comparisons without the added LayerNorm can be found in Supplement.~\ref{additional_throughput}. By default, we set autocast to false and evaluated FastVim, Vim, and ViT (without FlashAttention~\cite{dao2023flashattention, dao2022flashattention}) using float32 precision in line with VMamba~\cite{vmamba}. Additional analysis on larger-sized models and the impact of enabling or disabling autocast is available in Supplement.~\ref{additional_throughput}.

\subsection{Self-Supervised Learning: MAE}
\label{subexp:mae}

\noindent \textbf{Setting.} We extend the training paradigm of our FastVim models using self-supervision without labels. Specifically, we explore the Mask Autoencoder~\cite{mae} (MAE) approach, commonly used for self-supervising vision transformers, in the context of our proposed FastMaskVim method. We adhered to the same pre-training and similar fine-tuning settings as MAE's GitHub repository, with further details in the Supplement.~\ref{additional_mae_section}. For fine-tuning, we applied a drop path rate of 0.3 across all models, and found gradient clipping at 3 necessary for images sized \(448\). In MAE training, FastMaskVim serves as the encoder, while a lightweight decoder with Vim at a default depth of 2 and dimension of 512 was used. We also pre-trained Vim (base and large, both with layer norm for stability, see fig.~\ref{fig:stability}) with MAE to establish baselines. All models were trained for 1600 epochs with a masking ratio of 0.75.

\begin{table}[ht]
    \caption{Comparison of FastMaskVim with Vim and ViT, btoh pretrained with MAE, and other pretrained Vim based baselines ARM~\cite{arm} and HybridMH~\cite{hybridmh}. All models pre-trained with 224 image size on ImageNet-1k, and then end-to-end fine-tuned on 224 size image unless otherwise specified. B refers to Base, L to Large, H to Huge size models.}
    \vspace{-8pt}
    \begin{center}
    \resizebox{0.8\columnwidth}{!}{
    \begin{tabular}{lcccc}
        \toprule
        Model & B & L & H & H$_{448}$ \\
        \midrule
        ViT~\cite{mae} & 83.6 & 85.9 & 86.9 & 87.8 \\
        \midrule
        Vim & 83.3 & 85.1 & -- & -- \\
        ARM~\cite{arm} & 83.2 & 84.5 & 85.0 & -- \\
        HybridMH~\cite{hybridmh} & 84.9 & 85.0 & -- & -- \\
        \midrule
        \rowcolor[HTML]{E4E8FF} 
        FastMaskVim & 83.0 & 84.9 & 86.1 & 86.7 \\
        \bottomrule    
    \end{tabular}
}
\end{center}
\vspace{-4pt}
    \label{tab:mae_results}
\end{table}

\noindent \textbf{Results.} As illustrated in Table~\ref{tab:mae_results}, our proposed FastMaskVim, pretrained with MAE, performs on par with the Vim baseline with a minimal drop in performance of 0.3$\%$ and 0.2$\%$ across base and large model sizes respectively while being faster in all settings: pre-training, fine-tuning, and inference. Recently, the pure Mamba-based model ARM~\cite{arm} and the hybrid Mamba-based model HybridMH~\cite{hybridmh} have demonstrated state-of-the-art performance with autoregressive pretraining and masked autoregressive pretraining on ImageNet-1k, respectively. Here, we show that with minimal adjustments in the finetuning setup (further detailed in the Supplement~\ref{additional_mae_section}) and the application of post-SSM LayerNorm, MAE pretrained Vim and our FastVim can achieve comparable performance. Lastly, we demonstrate the scalability of FastMaskVim with images sized at 448, establishing a new state-of-the-art performance for Mamba-based methods in vision. We acknowledge that ViT excels in pretraining with MAE when compared with Vim, unlike in supervised training (Table~\ref{tab:additional_main_results}), highlighting the need for more exploration of MAE pre-training/fine-tuning recipe for the Vision Mamba. 

\subsection{Cell imaging: JUMP-CP}
\label{subexp:cell}

\noindent \textbf{Settings.} The JUMP-CP benchmark~\cite{chandrasekaran2023jump} serves as a microscopy imaging standard. The dataset includes a 160 perturbation classification task. We concentrated on the BR00116991 plate, containing 127k training, 45k validation, and 45k testing images. Each image has 8 channels: 5 for fluorescence and 3 for brightfield data. By default we keep channel-first scanning path (refer to Sec.~\ref{subsec:channelmodeling}) along with sorted HCS for both ChannelVim and FastChannelVim. Further ablations and implementation details can be found in Supplement.~\ref{additional_jumpcp_implementation_section}.

\begin{table}[ht]
    \caption{Benchmarks of 160-way perturbed gene prediction on JUMP-CP dataset. All methods use hierarchical channel sampling~\cite{channelvit} for training, and testing is done using all 8 channels. Each cell image is of resolution $224 \times 224 \times 8$}
    \vspace{-6pt}
    \begin{center}
    \resizebox{0.7\columnwidth}{!}{
    \begin{tabular}{lcc}
        \toprule

        Method & Token Grid & Top-1 \\
        &  &  (\%) \\

        \specialrule{1pt}{1pt}{1pt}

         ViT-S/16 & $14^2$  & 58.9 \\
         Vim-S/16 & $14^2$  & 61.0 \\
        \midrule
         ChannelViT-S/16 & $14^2\times8$  & 68.6 \\
         ChannelVim-S/16 & $14^2\times8$  & 73.5 \\
         \rowcolor[HTML]{E4E8FF} 
         FastCha.Vim-S/16 & $14^2\times8$  & 73.6 \\
        
        \specialrule{1pt}{1pt}{1pt}

         ViT-S/8 & $28^2$  & 67.6 \\
         Vim-S/8 & $28^2$  & 66.4 \\

        \midrule
         ChannelViT-S/8 & $28^2\times8$& 74.8 \\
         ChannelVim-S/8 & $28^2\times8$ & 83.0 \\
         \rowcolor[HTML]{E4E8FF} 
         FastCha.Vim-S/8 & $28^2\times8$  & 83.1 \\
            \bottomrule   

    \end{tabular}
}
\end{center}
\vspace{-6pt}
    \label{tab:jumpcp_results}
\end{table}

\noindent \textbf{Results.} In Table~\ref{tab:jumpcp_results}, we present our findings on the 160-way classification task. 
Due to the highly complementary nature of channel information, motivated by ChannelViT~\cite{channelvit}, we focus here on our proposed methods: ChannelVim and FastChannelVim. We use patch sizes of $16 \times 16$ and a high-resolution model with a patch size of $8 \times 8$. 

We observe that without per-channel tokenization, the Vim method performs on par with the ViT model for patch sizes 16 and 8. However, when using large token sequences with per-channel tokenization ($\times 8$ tokens), ChannelVim significantly surpasses ChannelViT by approximately 5\% at a patch size of 16, and this advantage grows to 8\% with even longer token sequences at a patch size of 8. 
These phenomenal improvements over the current standard transformer architectures underscore the necessity of replacing the Transformer backbone with the Mamba backbone for visual encoders in per-channel tokenization paradigms, such as microscopy cell imaging which has implications in the field of drug discovery~\cite{kenyon2024vitally, pham2024enhancing}. Additionally, it is evident that FastChannelVim-S maintains similar performance to the full-contextualization method ChannelVim-S, while offering significant speedup. From this observation, we believe patch size can be decreased further to get even more performance boost insipred by per-pixel tokenization~\cite{nguyen2024image}, which is extremely efficient with FastChannelVim and its extensions when compared to ChannelVim and ChannelViT.

\subsection{Semantic Segmentation}
\label{subexp:segmentation}

\noindent \textbf{Settings.} Here we conduct experiments on the ADE20K~\cite{ade20k} dataset using UperNet~\cite{upernet} as a segmentation framework for all backbones. The dataset contains 150 fine-grained semantic categories, with 20K, 2K, and 3K images for the train, validation, and test splits, respectively. Further settings can be found in Supplement.~\ref{additional_semantic_implementation_section}.

\begin{table}[ht]
    \caption{Semantic segmentation benchmarks on \textbf{ADE20K}~\cite{ade20k} dataset. UperNet~\cite{upernet} framework is used for all comparison backbones, with a crop size of 512 $\times$ 512.}
   \vspace{-12pt}
    \begin{center}
    \resizebox{0.4\columnwidth}{!}{
    \begin{tabular}{lcc}
        \toprule
        Backbone  &  mIoU \\

        \midrule
        DeiT-T  & 39.2 \\
         DeiT-S + MLN  & 43.8 \\
         DeiT-B + MLN  & 45.5 \\
        \midrule
        Vim-T  & 41.0 \\
         Vim-S  & 44.9 \\
         \midrule

        \rowcolor[HTML]{E4E8FF} 
        FastVim-T & 41.8 \\
        \rowcolor[HTML]{E4E8FF} 
        FastVim-S & 44.9 \\
        \rowcolor[HTML]{E4E8FF} 
        FastVim-B & 47.8 \\
        \bottomrule
        
    \end{tabular}
}
\end{center}
\vspace{-12pt}
    \label{tab:segmentation_results}
\end{table}

\noindent \textbf{Results.} As shown in Table~\ref{tab:segmentation_results}, FastVim consistently outperforms DeiT while achieving performance on par with Vim. The main aim of this study is to accelerate visual processing for larger images while ensuring competitive results. 
Previously, Vim~\cite{vim} demonstrated its significant advantages over DeiT in terms of GPU memory efficiency and speed as resolution increases. Our development further extends these benefits by creating an even faster yet capable Vision Mamba encoder: FastVim.

\subsection{Object Detection and Instance Segmentation}
\label{subexp:detection}

\noindent \textbf{Settings.} Here we conduct experiments on the MSCOCO 2017 dataset~\cite{coco} using Cascade Mask R-CNN with ViTDet~\cite{vitdet} for all backbones in line with Vim~\cite{vim}. The dataset contains 118K, 5K, and 20K images for training, validation, and testing, respectively. Further settings can be found in Supplement.~\ref{additional_objectdet_implementation_section}.

\begin{table}[ht]
\caption{Object detection and instance segmentation benchmarks on COCO dataset using Cascaded Mask R-CNN~\cite{mask_rcnn} framework. $^*$detection transfer conducted using provided Vim-S (GitHub).}
\vspace{-6pt}
    \begin{center}
    \resizebox{\columnwidth}{!}{
\begin{tabular}{l|ccc|ccc}
\toprule
Backbone & AP\textsuperscript{box} & AP\textsuperscript{box}$_{50}$ & AP\textsuperscript{box}$_{75}$ & AP\textsuperscript{mask} & AP\textsuperscript{mask}$_{50}$ & AP\textsuperscript{mask}$_{75}$ \\
\midrule
DeiT-T & 44.4 & 63.0 & 47.8 & 38.1 & 59.9 & 40.5  \\

\midrule
Vim-T & 45.7 & 63.9 & 49.6 & 39.2 & 60.9 & 41.7 \\
Vim-S$^*$ & 47.1 & 65.8 & 50.7 & 40.6 & 62.9 & 43.5 \\
\midrule
\rowcolor[HTML]{E4E8FF} 
FastVim-T & 45.1 & 63.7 & 48.5 & 39.0 & 60.8 & 41.6 \\
\rowcolor[HTML]{E4E8FF} 
FastVim-S & 48.4 & 67.2 & 52.2 & 41.8 & 64.3 & 44.7 \\
\rowcolor[HTML]{E4E8FF} 
FastVim-B & 50.0 & 68.7 & 54.2 & 43.2 & 66.0 & 46.6 \\

\bottomrule
\end{tabular}
}
\end{center}
\vspace{-6pt}
    \label{tab:detection_results}

\end{table}

\noindent \textbf{Results.} In Table~\ref{tab:detection_results}, we see our FastVim-T performs comparably to Vim-T while surpassing Vim-S by 1.3 $\text{AP}^\text{box}$/1.2 $\text{AP}^\text{mask}$. These results highlight the effectiveness of our FastVim, even when handling larger 64×64 token grids in 1024×1024 MSCOCO images. Our method remains competitively performant despite pooling 64 tokens at once—significantly more than the 14 or 32 tokens pooled in ImageNet-1k and ADE20K—illustrating the scalability of our approach to higher resolutions.

\subsection{Ablation Study}
\label{subexp:ablation}

\textbf{Effect of Alternating Dimension Pooling on FastVim.} Here we investigate the importance of alternating spatial dimensions for pooling the token grid after each block in FastVim. As demonstrated in Table~\ref{tab:ablation_rotate}, FastVim with alternating pooling dimensions outperforms configurations that consistently pool tokens either across columns (Pool$_{col}$) or rows (Pool$_{row}$) across all blocks. This suggests that alternating pooling dimensions facilitates more effective sparse communication between tokens.

\begin{table}[!h]
    \caption{Effect of alternating dimension pooling on ImageNet-1k.} 
    \vspace{-15pt}
    \begin{center}
    \resizebox{0.7\columnwidth}{!}{
    \begin{tabular}{lccc}
        \toprule
        Model & FastVim-S & Pool$_{col}$ & Pool$_{row}$ \\
        \midrule
        Top-1 ($\%$)  & 81.1 & 80.0 & 79.6 \\ 
        \bottomrule
    \end{tabular}
}
\end{center}
\vspace{-14pt}
    \label{tab:ablation_rotate}
\end{table}

\noindent \textbf{Exploring pooling in ViT.} We now examine how our pooling method performs with the contextualization module in Transformers, namely Self-Attention. As seen in Table~\ref{tab:ablation_vit}, our method, which applies alternating Pool$_{col}$ and Pool$_{row}$ pooling across blocks, performs significantly worse compared to the baseline ViT-S, which was trained using the default settings from DeiT\cite{deit}. It can be argued that Mamba has a 1D convolution (conv1d) layer which can do contextualization of tokens to an extent. To address that we trained 2 variations, FastVim without conv1d layers and ViT-S with Pool and conv1d for fair comparison. We can observe that conv1d is particularly helpful only in the Mamba case and ViT can't benefit from our proposed pooling method. This outcome highlights that our proposed approach is particularly well-suited to the emerging Mamba architecture. We note that while the failure of pooling approach in retaining performance in ViT as compared to Vim is interesting, it has opportunities for further exploration.

\begin{table}[ht]
    \caption{Effect of pooling in ViT on ImageNet-1k.} 
    \vspace{-15pt}
    \begin{center}
    \resizebox{\columnwidth}{!}{
    \begin{tabular}{lccccc}
        \toprule
        Model & ViT-S & w/ Pool & w/ Pool w/ conv1d & FastVim-S &  w/o conv1d \\
        \midrule
        Top-1 (\%)  & 80.1 & 73.9 & 74.0 & 81.1 & 78.4 \\ 
        \bottomrule
    \end{tabular}
}
\end{center}
\vspace{-14pt}
    \label{tab:ablation_vit}
\end{table}

\noindent \textbf{Additional Ablations:} In Supplement.~\ref{additional_ablations} and~\ref{additional_jumpcp_implementation_section}, we additionally explore 1) the effect of using a class token in FastViM, 2) the performance impact of different input norm and post-ssm norm combinations such as RMS-LN (default in FastViM), RMS-RMS, and LN-LN, 3) Effect of decompression after the skip connection ($\mathbf{D}$ in fig.~\ref{fig:model}), and 4) comparisons between mean, max, and attention pooling in FastVim.

\vspace{-0.1cm}
\section{Related work}
\label{sec:related work}

\noindent \textbf{Vision Mamba.} VMamba~\cite{vmamba} introduced visual state space blocks that combine Mamba with 2D convolution layers and a hierarchical design similar to the Swin transformer~\cite{liu2021swin}, employing tricks like reducing the SSM states and expansion ratio to optimize throughput. EfficientVMamba~\cite{efficientvmamba} enhances VMamba by using an atrous-based selective scanning strategy for efficient global feature extraction, integrating SSMs with convolution branches. GroupMamba~\cite{groupmamba} addresses scalability and stability with a Modulated Group Mamba layer featuring multi-directional scanning and enhanced cross-channel communication. MambaVision~\cite{mambavision} reconfigures Mamba by incorporating convolutional layers and interleaved Mamba-Transformer blocks, achieving a new state-of-the-art in accuracy and throughput. Our average pooling in FastVim can be readily applied to these advancements.

\noindent \textbf{Sparse contextualization methods.} Inspired by efforts to enhance efficiency in ViTs, numerous studies~\cite{ryoo2021tokenlearner, hou2022token, bolya2022token, renggli2022learning, liang2022not, rao2021dynamicvit} have examined the reduction of tokens across layers through merging or pruning. Similarly, Famba-V~\cite{famba} utilizes a token fusion technique to consolidate similar tokens in Vim, thereby reducing training and inference time. Vim-prune~\cite{zhan2024exploring, zhan2024rethinking} addresses the challenges of naive pruning in Mamba due to its sequential nature by introducing pruning-aware hidden state alignment to stabilize neighborhoods. Our FastVim method aligns with this sparse contextualization approach, offering simplicity and maintaining performance beyond the tiny model size, in contrast to Famba and Vim-prune. Additionally, our proposed extension, FastMaskVim, can be seamlessly integrated with Vim-prune or Famba for further speedup, albeit with potential performance trade-offs. 

\vspace{-0.1cm}

\section{Conclusion and Future Work}
\label{sec:conclusion}

We presented FastVim, which enhances Vim's efficiency by reducing its computational complexity and increasing practical throughput speed. Remarkably, FastVim achieves this without any performance degradation compared to the baseline Vim model across multiple tasks, even though it contextualizes significantly fewer tokens in the SSM scan at each layer. By using pooling, our method delivers up to 72.5\% overall throughput speedup (while reducing the parallel steps in SSM scan by \(2\times\)), with the gap widening at higher resolutions (longer token sequences). Our FastMaskVim sets the new state-of-the-art performance of \textit{86.7\%} on ImageNet-1k for Mamba-based encoders and ranks among the top 12 visual encoders (when only ImageNet-1k is used). Additionally, it achieves substantial improvements over Transformer baselines in microscopy imaging. Beyond extending to Mamba-2~\cite{mamba2}, future work will also explore applying FastVim in gigapixel imaging, such as histopathology~\cite{nasiri2024vim4path, graikos2024learned, xu2024whole}, as well as in video domain~\cite{li2025videomamba, das2021vpn++, ashutosh2023hiervl} aligning with FastChannelVim.

\vspace{-0.2cm}

\section{Acknowledgments}
\vspace{-0.1cm}

We thank Emily Fox for the great help in preparing the manuscript, Angela Oliveira Pisco and Tommaso Dreossi for thought-provoking discussions, and Patrick Conrad for GPU infrastructure support. 
 
{
    \small
    \bibliographystyle{ieeenat_fullname}
    \bibliography{main}
}

\clearpage
\maketitlesupplementary

\definecolor{lightblue}{RGB}{173, 216, 255}
\begin{table}[ht]
    \caption{Classification benchmarks on \textbf{ImageNet-1k}~\cite{imagenet} dataset. All models are trained from scratch on Image size of $224\times224$. $\dag$ denotes we extend the training of Vim to base-size model.} 
    \begin{center}
    \resizebox{0.85\columnwidth}{!}{
    \begin{tabular}{lccc}
        \toprule
        Model & \#Params &  FLOPs & Top-1 \\

         & (M) & (G) & (\%) \\
        
        \midrule
        \multicolumn{4}{c}{Conv-Based} \\
        \midrule

        ConvNeXt-T~\cite{liu2022convnet} & 29M & 4.5G & 82.1 \\
        ConvNeXt-S~\cite{liu2022convnet} & 50M & 8.7G & 83.1 \\
        ConvNeXt-B~\cite{liu2022convnet} & 89M & 15.4G & 83.8 \\
        \midrule
        \multicolumn{4}{c}{Transformer-Based} \\
        \midrule
        DeiT-T~\cite{touvron2021training} & 6M & 1.3G & 72.2 \\
        DeiT-S~\cite{touvron2021training} & 22M & 4.6G & 79.8 \\
        DeiT-B~\cite{touvron2021training} & 86M & 17.5G & 81.8 \\

        Swin-T~\cite{liu2021swin} & 28M & 4.5G & 81.3 \\
        Swin-S~\cite{liu2021swin} & 50M & 8.7G & 83.2 \\
        Swin-B~\cite{liu2021swin} & 88M & 15.4G & 83.5 \\

        \midrule
        \multicolumn{4}{c}{Hybrid (Mamba + \{2D convolution, Attention module\}) } \\
        \midrule

        VMamba-T~\cite{vmamba} & 31M & 4.9G & 82.5 \\
        VMamba-S~\cite{vmamba} & 50M & 8.7G & 83.6 \\
        VMamba-B~\cite{vmamba} & 89M & 15.4G & 83.9 \\

        Eff.VMamba-T~\cite{efficientvmamba} & 6M & 0.8G & 76.5 \\
        Eff.VMamba-S~\cite{efficientvmamba} & 11M & 1.3G & 78.7 \\
        Eff.VMamba-B~\cite{efficientvmamba} & 33M & 4.0G & 81.8 \\

        MambaVision-T~\cite{mambavision} & 32M & 4.4G & 82.3 \\
        MambaVision-S~\cite{mambavision} & 50M & 7.5G & 83.3 \\
        MambaVision-B~\cite{mambavision} & 98M & 15.0G & 84.2 \\

        \midrule
        \multicolumn{4}{c}{Pure Mamba architecture} \\
        \midrule

        Vim-T~\cite{vim} & 7M & 1.8G & 76.1 \\
        Vim-S~\cite{vim} & 26M & 5.9G & 80.5 \\
        

        Vim-B~\cite{vim}$\dag$ & 98M & 20.9G & 80.7 \\
        Vim-B w/ LN$\dag$ & 98M & 21.0G & 82.6 \\

        PlainMamba-L1~\cite{plainmamba} & 7M & 3.0G & 77.9 \\
        PlainMamba-L2~\cite{plainmamba} & 25M & 8.1G & 81.6 \\
        PlainMamba-L3~\cite{plainmamba} & 50M & 14.4G & 82.3 \\
        
        Mamba\textsuperscript{\textregistered}-T~\cite{wang2024mamba} & 9M & 1.9G & 77.4 \\
        Mamba\textsuperscript{\textregistered}-S~\cite{wang2024mamba} & 28M & 6.3G & 81.1 \\
        Mamba\textsuperscript{\textregistered}-B~\cite{wang2024mamba} & 99M & 22.1G & 82.9 \\

        \rowcolor[HTML]{E4E8FF}  
        FastVim-T & 7M & 1.17 & 75.4 \\
        
        \rowcolor[HTML]{E4E8FF}  
        FastVim-S & 26M & 4.43 & 81.1 \\
        
        \rowcolor[HTML]{E4E8FF}  
        FastVim-B & 98M & 17.23 & 82.6 \\

        \bottomrule
    \end{tabular}
}
\end{center}
    \label{tab:additional_main_results}
\end{table}

In this supplementary material,  details are provided on the following: 
\begin{itemize}

    \item Self-Supervised Learning: MAE (additional) (\ref{additional_mae_section})

    \item Additional ablations 
    (\ref{additional_ablations})

    \item JUMP-CP (additional)  
    (\ref{additional_jumpcp_implementation_section})

    \item Additional Throughput analysis
    (\ref{additional_throughput})

    \item Semantic Segmentation implementation details  
    (\ref{additional_semantic_implementation_section})

    \item Object Detection and Instance Segmentation implementation details  
    (\ref{additional_objectdet_implementation_section})

    \item Kernel details
    (\ref{kernel_details})

    \item Model configurations
    (\ref{model_sizes})

\end{itemize}

\section{Self-Supervised Learning: MAE (additional)}
\label{additional_mae_section}

\noindent \textbf{Implementation Details}. We closely followed the pre-training (Table~\ref{tab:mae_pretraining}), fine-tuning (Table~\ref{tab:mae_finetuning}), and linear-probing (Table~\ref{tab:mae_linearprobe}) settings from the Masked Autoencoders~\cite{mae} codebase. All MAE pretraining is done for 1600 epochs in this study. A few key changes, particularly for fine-tuning and linear probing, are discussed below.

\begin{table}[h]
    \caption{MAE: Pre-training setting.} 
    \begin{center}
    \resizebox{0.8\columnwidth}{!}{
    \begin{tabular}{l|l}
        \toprule
        config & value \\
        \midrule
optimizer                     & AdamW~\cite{adamw}    \\ 
base learning rate            & 1.5e-4  \\ 
weight decay                  & 0.05    \\ 
optimizer momentum            & $\beta_1, \beta_2=0.9, 0.95$    \\ 
batch size                    & 4096    \\ 
learning rate schedule        & cosine decay~\cite{loshchilov2016sgdr}  \\ 
training epochs    & 1600    \\ 
warmup epochs    & 40    \\ 
augmentation                  & RandomResizedCrop        \\      
        \bottomrule
    \end{tabular}
}
\end{center}
    \label{tab:mae_pretraining}
\end{table}

\begin{table}[h]
    \caption{MAE: End-to-end fine-tuning setting. Note that layer-wise lr decay is applied after every two blocks instead of one.} 
    \begin{center}
    \resizebox{0.8\columnwidth}{!}{
    \begin{tabular}{l|l}
        \toprule
        config & value \\
        \midrule
optimizer                & AdamW                                             \\ 
base learning rate       & 5e-4 (B), 1e-3 (L/H)                                             \\ 
weight decay             & 0.05                                              \\ 
optimizer momentum       & $\beta_1$, $\beta_2$=0.9, 0.999                   \\ 
layer-wise lr decay~\cite{bao2021beit} & 0.65 (B), 0.75  (L/H)                     \\ 
batch size               & 1024                                              \\ 
learning rate schedule   & cosine decay                                      \\ 
warmup epochs            & 5                                                 \\ 
training epochs          & 100 (B), 50 (L/H)                                 \\ 
augmentation             & RandAug (9, 0.5)~\cite{cubuk2020randaugment}                    \\ 
label smoothing & 0.1                                         \\ 
mixup~\cite{zhang2017mixup}       & 0.8                                               \\ 
cutmix~\cite{yun2019cutmix}      & 1.0                                               \\ 
drop path~\cite{huang2016deep}   & \textbf{0.3}                                 \\

        \bottomrule
    \end{tabular}
}
\end{center}
    \label{tab:mae_finetuning}
\end{table}

\begin{table}[h]
    \caption{MAE: Linear probing setting.} 
    \begin{center}
    \resizebox{0.8\columnwidth}{!}{
    \begin{tabular}{l|l}
        \toprule
        config & value \\
        \midrule
optimizer                & SGD                                             \\ 
base learning rate       & 0.1                                             \\ 
weight decay             & 0                                             \\ 
optimizer momentum       & 0.9                    \\ 
batch size               & 4096                                              \\ 
learning rate schedule   & cosine decay                                      \\ 
warmup epochs            & 10                                                 \\ 
training epochs          & 90                                 \\ 
augmentation             & RandomResizedCrop                    \\ 

        \bottomrule
    \end{tabular}
}
\end{center}
    \label{tab:mae_linearprobe}
\end{table}

\noindent \textbf{Key Recipe Details.} 

1) Since Vim contains twice the number of layers compared to ViTs, we decreased the layer-wise learning rate decay every two blocks, instead of every block as in the MAE codebase for ViT fine-tuning, to ensure adequate fine-tuning of the initial layers. 2) We applied a scaling factor of \(1 - \text{mask ratio}\) (75\% masking by default) during fine-tuning and linear probing when pooling tokens before the SSM block. During pretraining, each row averaged 25\% of the tokens. In FastMaskVim, we sum the unmasked tokens and then divide by the number of columns for pooling instead of using mean pooling. To align this in fine-tuning and linear probing tasks, a scaling factor of 0.25 was necessary to achieve better performance.
\\

\noindent \textbf{Ablations}.

\begin{enumerate}

\item \textbf{Divide by number of columns vs. mean pool in FastMaskVim.} In Table~\ref{tab:ablation_fastmae_constantdivide}, we compare the performance of pre-training FastMaskVim using the default setting, where the sum of tokens in a row is divided by the number of columns, against mean pooling, where each row's sum is divided by the number of unmasked tokens present in the row. We observe in MAE pre-training that mean pooling performs slightly worse compared to the constant divide technique in corresponding downstream fine-tuning. However, exploring the comparison between mean pooling and constant divide in the context of supervised training is left for future research.

\begin{table}[!h]
    \caption{Comparison of constant divide vs. mean pool in pre-training FastMaskVim} 
    \begin{center}
    \resizebox{1\columnwidth}{!}{
    \begin{tabular}{lcc}
        \toprule
        FastMaskVim-B & Constant divide (default) & Mean Pool \\
        \midrule
        Top-1 ($\%$)  & 83.0 &  82.8 \\ 
        \bottomrule
    \end{tabular}
}
\end{center}
    \label{tab:ablation_fastmae_constantdivide}
\end{table}

\item \textbf{Finetuning with alternate layer lr decay.} In Table~\ref{tab:ablation_fastmae_lrdecay}, we compare the performance of fine-tuning pre-trained FastMaskVim using alternate layer learning rate decay instead of per-layer decay as in the MAE codebase. We observe that, since Vim typically contains twice the number of layers compared to ViTs with a similar parameter count, adjusting the decay logic to apply the learning rate decay every 2 blocks was necessary to ensure adequate fine-tuning of the early layers. With this simple adjustment, we were able to improve performance by a significant margin of 1\%. This analysis motivates us to believe that with further recipe improvements, FastVim can match the performance of ViTs with MAE pre-training, where we currently observe a lag of 0.6-1\% across Base to Huge model sizes (see Table~\ref{tab:mae_results}).

\begin{table}[!h]
    \caption{Comparison of alternate layer lr decay vs. per layer lr decay in finetuning} 
    \begin{center}
    \resizebox{1\columnwidth}{!}{
    \begin{tabular}{lcc}
        \toprule
        FastMaskVim-L & Alternate lr decay (default) & All layer lr decay \\
        \midrule
        Top-1 ($\%$)  & 84.9 &  83.9 \\ 
        \bottomrule
    \end{tabular}
}
\end{center}
    \label{tab:ablation_fastmae_lrdecay}
\end{table}

\item \textbf{Finetuning with scaling factor.} In Table~\ref{tab:ablation_fastmae_finetune}, we demonstrate the effect of using a scaling factor (0.25) in the fine-tune transfer of pre-trained FastMaskVim. Applying the scaling factor results in an improvement of 0.3\% compared to the default mean pooling in fine-tuning without multiplying by the scaling factor. As shown in Fig.~\ref{fig:finetune_scaling}, when scaling is not used, the initial performance is much lower, although it catches up closely by the end of the training schedule.

\begin{table}[!h]
    \caption{Effect of scaling factor in finetuning} 
    \begin{center}
    \resizebox{0.8\columnwidth}{!}{
    \begin{tabular}{lcc}
        \toprule
        FastMaskVim-L & w/ scaling (default) & w/o scaling \\
        \midrule
        Top-1 ($\%$)  & 84.9 &  84.6 \\ 
        \bottomrule
    \end{tabular}
}
\end{center}
    \label{tab:ablation_fastmae_finetune}
\end{table}

\begin{figure}[h]
\centering
    \includegraphics[width=0.8\linewidth]{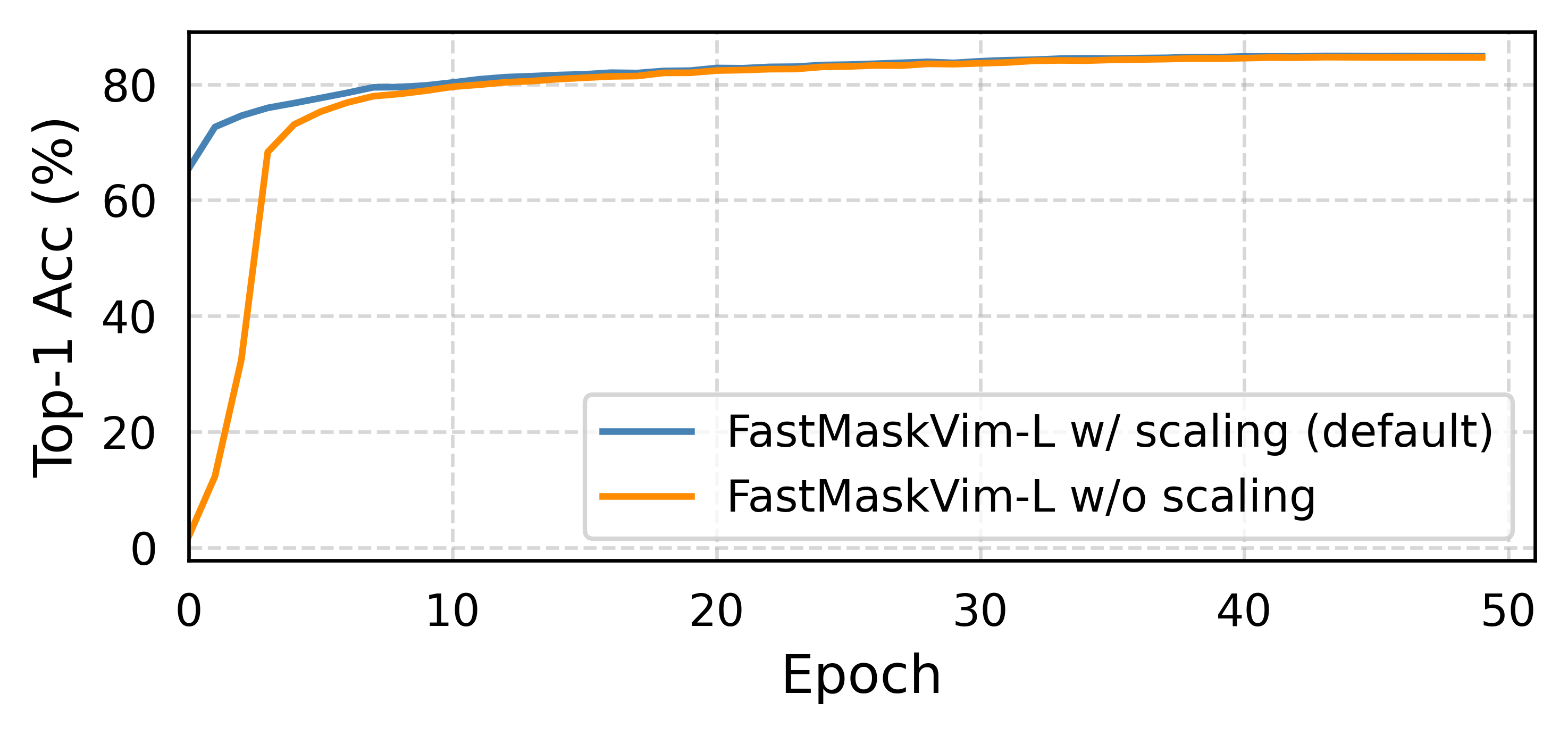}
    \caption{Effect of scaling factor in finetuning performance from MAE pretraining FastMaskVim-L on ImageNet-1k.
    }
    \label{fig:finetune_scaling}
\end{figure}

\item \textbf{Linear probing with scaling factor.} In Table~\ref{tab:ablation_fastmae_linearprobe}, we compare the linear probing performance of FastMaskVim with and without the scaling factor (0.25). We observe a drastic difference in performance and note that without the scaling factor, the model was unable to train due to the significant difference between the pre-training and linear probing distributions of number of unmasked tokens. During pre-training, on average, each row had 25\% of the number of columns (or number of rows when transposed in alternate layers) as unmasked tokens. Since we divided by the number of columns following the sum operation, the signal magnitude was in a lower range. In contrast, during linear probing, because all tokens are unmasked, we add the number of column tokens and divide by the number of columns, resulting in a very different signal range. We further compared the performance with pre-trained Vim's performance in linear probing and found that it performs considerably worse than FastVim. In MAE pre-training, random masking disrupts the sequential token positions, as demonstrated by Vim-prune~\cite{zhan2024exploring}. In contrast, during linear probing, this issue does not exist, causing a shift in the neighborhood distribution and resulting in low linear probe performance. This issue does not occur in our FastVim since, in both pre-training and linear probing, the number of rows/columns remains the same, ensuring that the neighborhood remains consistent. 

\begin{table}[!h]
    \caption{Effect of scaling factor in linear probing} 
    \begin{center}
    \resizebox{1\columnwidth}{!}{
    \begin{tabular}{lccc}
        \toprule
        Method & Vim-L & FastMaskVim-L w/ scaling & w/o scaling \\
        \midrule
        Top-1 ($\%$) & 45.6 & 60.2 &  0.02 \\ 
        \bottomrule
    \end{tabular}
}
\end{center}
    \label{tab:ablation_fastmae_linearprobe}
\end{table}

\end{enumerate}

\section{Additional ablations}
\label{additional_ablations}

\begin{enumerate}

\item \textbf{Effect of using class token in FastVim.} In Table~\ref{tab:ablation_classtoken}, we compare the performance of FastVim-S with a class token versus without a class token (default). We observe that having a class token improves performance but leads to slower reshape-transpose, pooling, and repeat operations to handle the middle class token. Since the goal of this study is to improve throughput while maintaining performance relative to the Vim baseline, we proceeded with all experiments without a class token. However, it is worth noting that even with a class token, our method is faster than Vim, according to our preliminary analysis. One future direction could be to mean pool only middle rows/columns for image-level representation instead of current mean pooling of all tokens for image representation. 
\\

\begin{table}[!h]
    \caption{Effect of using class token in FastVim on ImageNet-1K.} 
    \begin{center}
    \resizebox{0.8\columnwidth}{!}{
    \begin{tabular}{lcc}
        \toprule
        Model & FastVim-S & w/ Class token \\
        \midrule
        Top-1 ($\%$)  & 81.1 &  81.3 \\ 
        \bottomrule
    \end{tabular}
}
\end{center}
    \label{tab:ablation_classtoken}
\end{table}

\item  \textbf{The performance impact of different input norm and post-ssm norm combinations.} In Table~\ref{tab:ablation_norm_combinations}, we empirically demonstrate the performance of FastVim trained with different combinations of input normalization and post-SSM normalization. We found that using RMS normalization as the input norm and LayerNorm as the post-SSM norm yields the best performance.
\\

\begin{table}[!h]
    \caption{Effect of using different normalization combination in FastVim-S on ImageNet-1K.} 
    \begin{center}
    \resizebox{0.8\columnwidth}{!}{
    \begin{tabular}{lccc}
        \toprule
        Model & RMS-LN  & RMS-RMS & LN-LN \\
        \midrule
        Top-1 ($\%$)  & 81.1 & 80.7 & 80.9  \\ 
        \bottomrule
    \end{tabular}
}
\end{center}
    \label{tab:ablation_norm_combinations}
\end{table}

\item  \textbf{Effect of decompression after the skip connection on models' performance.} In Table~\ref{tab:ablation_decompress}, we explore whether in Fig.~\ref{fig:model}, we can move the skip connection \(\mathbf{D} x_t\) before repeating/decompressing the output to achieve even more speedup. However, we empirically found that it leads to a significant decrease in performance.
\\

\begin{table}[!h]
    \caption{Effect of decompression after the skip connection in FastVim-S on ImageNet-1K.} 
    \begin{center}
    \resizebox{0.8\columnwidth}{!}{
    \begin{tabular}{lcc}
        \toprule
        Model & Before $\mathbf{D}$ (default)  & After $\mathbf{D}$ \\
        \midrule
        Top-1 ($\%$)  & 81.1 & 78.7  \\ 
        \bottomrule
    \end{tabular}
}
\end{center}
    \label{tab:ablation_decompress}
\end{table}

\end{enumerate}

\section{JUMP-CP (additional)}
\label{additional_jumpcp_implementation_section}

\noindent \textbf{Implementation details}.

We followed the implementation details primarily from ChannelViT~\cite{channelvit}. Specifically, we used a learning rate of \(1 \times 10^{-3}\), a batch size of 256, and trained the model for 100 epochs, including 10 warmup epochs. We set the drop path rate to 0.05 and did not use EMA. All details and configuration files will be made available in the code.
\\

\noindent \textbf{Ablations}. 

\begin{enumerate}

\item  \textbf{ChannelVim-S: Effect of Spatial-First vs. Channel-First with and without sorted HCS.} In Table~\ref{tab:jumpcp_results_channelspatial}, we demonstrate the key configurations required to extend ChannelViT~\cite{channelvit} to the Mamba-based encoder, termed ChannelVim. As explained in detail in Sec.~\ref{subsec:channelmodeling}, due to the sequential processing in Mamba, the order of tokens matters. We found that the Channel-First method performs significantly better than Spatial-First. Whereas, the effect of sorting the output of hierarchical channel sampling (HCS) is opposite: it might be acting as an augmentation in the Spatial-First approach due to the order covering channel-by-channel, while it might be causing disruption in the neighborhood in the Channel-First approach since every next token in the sequence is another channel. Randomly shuffling the channel order (no sort) makes it difficult for learning.
\\

\begin{table}[!h]
    \caption{ChannelVim-S: Effect of Spatial-First vs. Channel-First with and without sorted HCS on 160-way perturbed gene prediction on JUMP-CP dataset. All methods use hierarchical channel sampling~\cite{channelvit} for training, and testing is done using all 8 channels. Each cell image is of resolution $224 \times 224 \times 8$.}
    \begin{center}
    \resizebox{0.7\columnwidth}{!}{
    \begin{tabular}{lcc}
        \toprule
        Method & HCS & Top-1 \\
        \midrule 
         Channel-First & Sorted  & 73.5 \\
          & Unsorted & 69.4 \\

        \midrule
         Spatial-First & Sorted & 65.9 \\
          & Unsorted & 67.9 \\
         
            \bottomrule    
    \end{tabular}
}
\end{center}
    \label{tab:jumpcp_results_channelspatial}
\end{table}

\item  \textbf{FastChannelVim-S: Effect of different pooling methods (mean, max, and attention pooling):} In this study, we use average pooling of tokens to compress the tokens before the SSM scan. We then explore the effect of different pooling methods, such as max pooling~\cite{ranasinghe2023perceptual} and attention pooling~\cite{abmil}, as detailed in Table~\ref{tab:jumpcp_results_meanmaxatt} on the JUMP-CP dataset. For attention pooling, we added a simple linear layer before each pooling layer to project each token to a size of one. This is followed by a SoftMax operation across tokens in the row, which is multiplied by a learned attention value and then summed across the row. We found that at a patch size of 16, all methods perform comparably. However, at a patch size of 8, max pooling and attention pooling methods start to perform better, likely due to the increased number of tokens in a row, allowing them to capture the most discriminative signals more effectively than mean pooling. Based on the accuracy-throughput trade-off, max pooling emerges as the best choice on the JUMP-CP dataset, as it is as fast as mean pooling while performing very close to attention pooling. Exploring effect of these pooling operation in natural imaging is left for future studies. 
\\

\begin{table}[!h]
    \caption{FastChannelVim-S: Effect of different pooling methods (mean, max, and attention pooling) on 160-way perturbed gene prediction on JUMP-CP dataset. All methods use hierarchical channel sampling~\cite{channelvit} for training, and testing is done using all 8 channels. Each cell image is of resolution $224 \times 224 \times 8$.}
    \begin{center}
    \resizebox{0.7\columnwidth}{!}{
    \begin{tabular}{lcc}
        \toprule
        Pooling & patch-size & Top-1 \\
        \midrule 
         Mean & 16  & 73.6 \\
          Max &  16 &  72.9 \\
          Att &  16 &  73.1 \\

        \midrule
        Mean & 8  & 83.1 \\
          Max &  8 &  85.0 \\
          Att &  8 &  85.8 \\

            \bottomrule    
    \end{tabular}
}
\end{center}
    \label{tab:jumpcp_results_meanmaxatt}
\end{table}

\item  \textbf{FastChannelVim-S: Effect of Pooling across 2 dimensions:} So far, we have explored pooling along only one spatial dimension, either across rows or columns. Now, we preliminarily explore pooling along two dimensions, which is particularly applicable in 3-dimensional datasets. When performing channel-wise tokenization, we obtain a 3D token grid. We experiment with the following pooling combinations in sequence every three blocks: column-channel pooling - row-channel pooling - row-column pooling - repeat. This approach provides much stronger compression, reducing the 3D token grid to a 1D token grid for the SSM scan.

\begin{table}[!h]
    \caption{FastChannelVim-S: Effect of Pooling across 2 dimensions on 160-way perturbed gene prediction on JUMP-CP dataset. All methods use hierarchical channel sampling~\cite{channelvit} for training, and testing is done using all 8 channels. Each cell image is of resolution $224 \times 224 \times 8$.}
    \begin{center}
    \resizebox{0.7\columnwidth}{!}{
    \begin{tabular}{lcc}
        \toprule
       Pooling & patch-size & Top-1 \\
        \midrule 
         Mean - 1D & 16  & 73.6 \\
          Max - 1D &  16 &  72.9 \\
         Mean - 2D & 16  & 74.3 \\
          Max - 2D &  16 &  73.5 \\
        \midrule
        Mean - 1D & 8  & 83.1 \\
          Max - 1D &  8 &  85.0 \\
        Mean - 2D & 8  & 78.4 \\
          Max - 2D &  8 &  84.0 \\
            \bottomrule    
    \end{tabular}
}
\end{center}
    \label{tab:jumpcp_results_2dpooling}
\end{table}

In Table~\ref{tab:jumpcp_results_2dpooling}, we demonstrate that at a patch size of 16 (token grid 14x14x8), both mean pooling and max pooling with 2D pooling work well and are on par with 1D pooling. In contrast, at a patch size of 8 (token grid 28x28x8), given the significantly larger number of tokens to pool (28x28 in row-column, 28x8 in row-channel, 28x8 in column-channel pooling blocks), mean pooling does not perform well. However, when we use max pooling, it performs much better, achieving results on par with ChannelVim with a patch size of 8 (see Table~\ref{tab:jumpcp_results}). Thus, even after pooling a much larger number of tokens, our method, FastChannelVim, still performs well with max pooling. This has implications in making the video models even faster~\cite{kahatapitiya2025object}.

\end{enumerate}

\section{Additional Throughput analysis}
\label{additional_throughput}

All throughput analysis is done on the H100 and involves inference throughput unless otherwise specified. 
\\

\begin{figure*}[!h]
    \centering
    \begin{subfigure}[b]{0.49\textwidth}
        \centering
        \includegraphics[width=\linewidth]{figures/speed_comparison_autocastfalse_h100.png}
        \caption{Autocast as False}
    \end{subfigure}
    \begin{subfigure}[b]{0.49\textwidth}
        \centering
        \includegraphics[width=\linewidth]{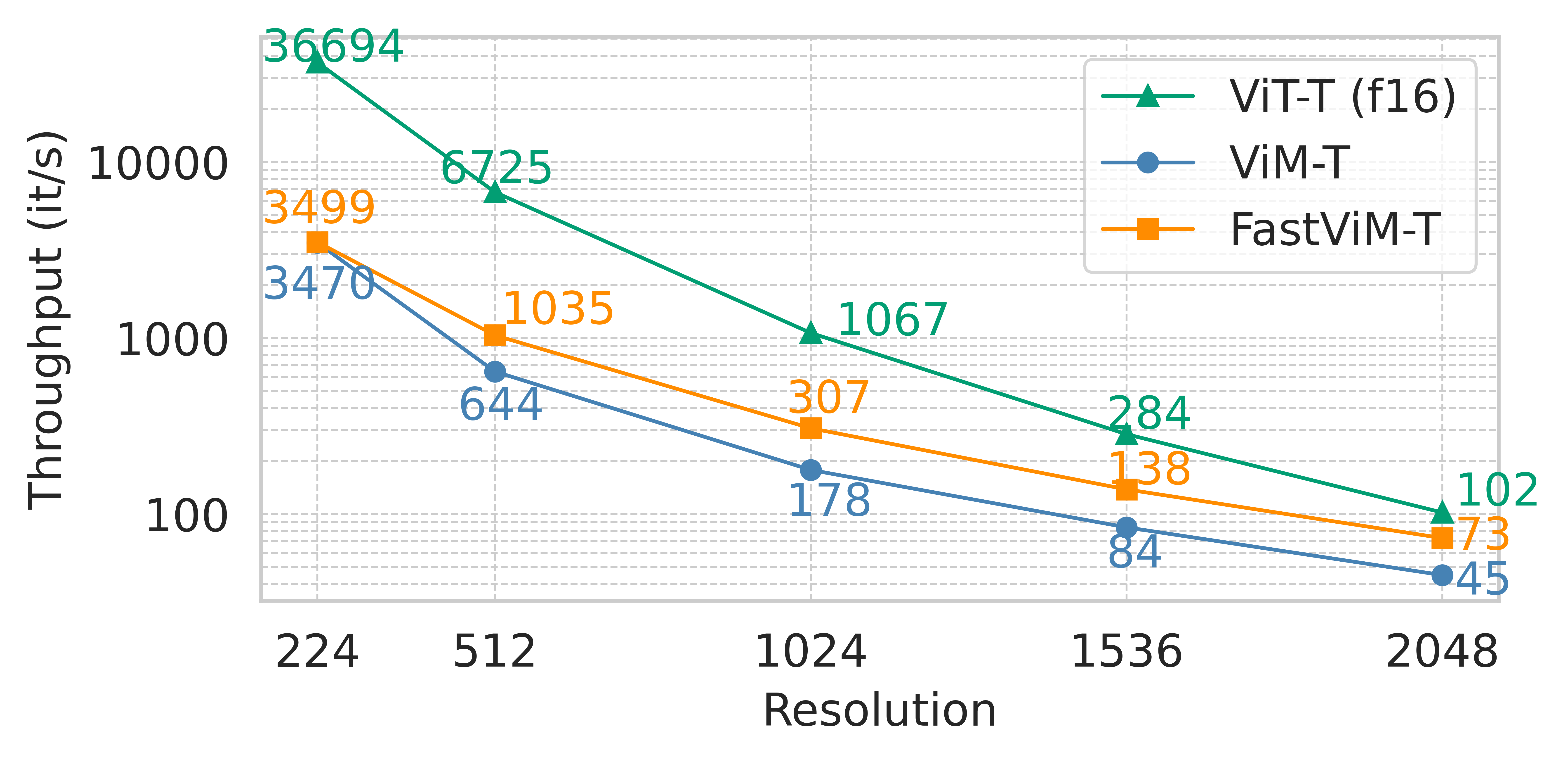}
        \caption{Autocast as True}
    \end{subfigure}
    \caption{Comparison of Inference Throughput (it/s) for FastVim, Vim, and ViT across different resolutions. Tested on H100 GPU with batch size of 128, and with LayerNorm post-SSM in Vim and FastVim.}
    \label{fig:throughput_autocast_on}
\end{figure*}

\begin{figure*}[!h]
    \centering
    \begin{subfigure}[b]{0.49\textwidth}
        \centering
        \includegraphics[width=\linewidth]{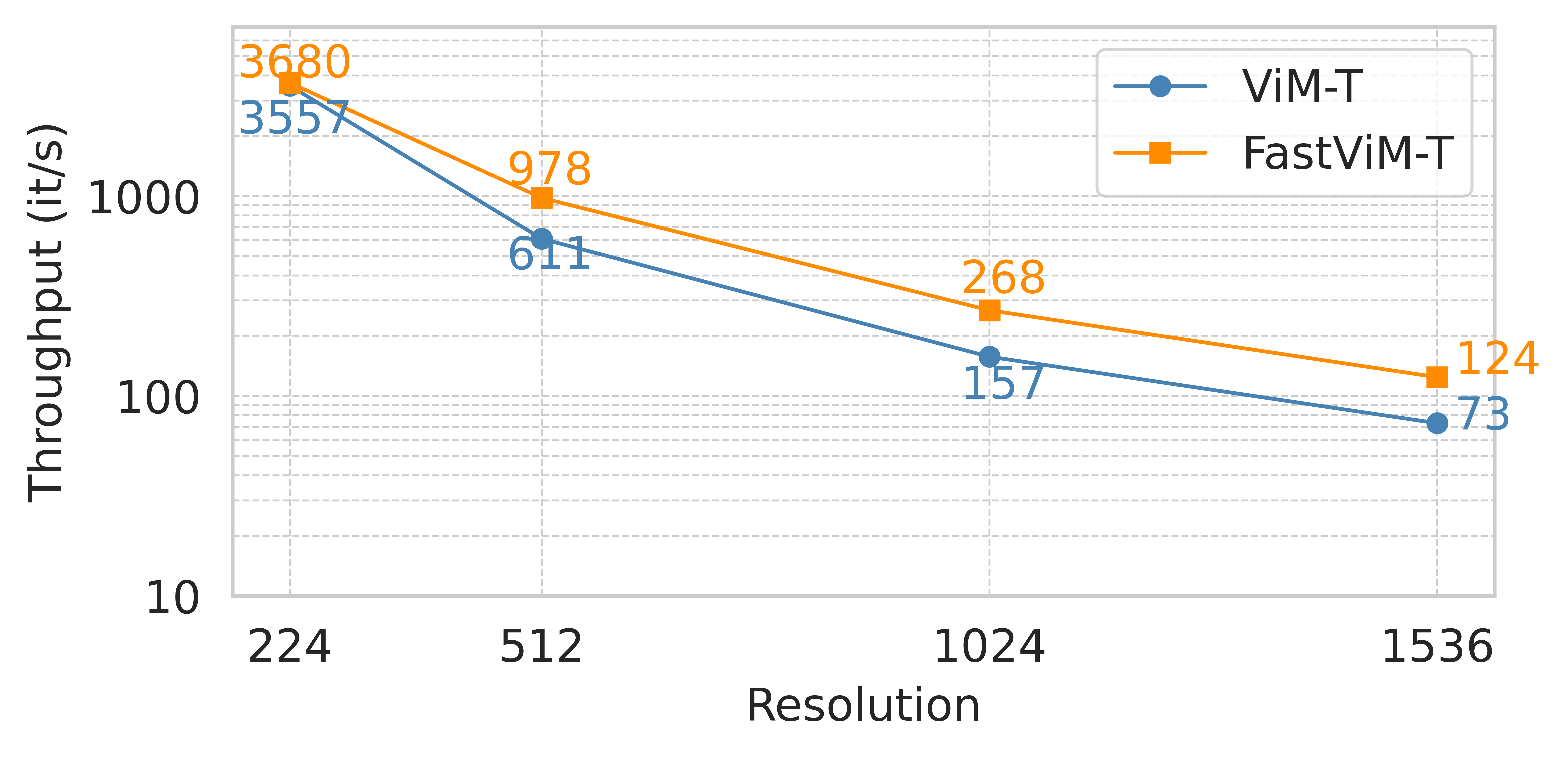}
        \caption{On H100 GPU}
    \end{subfigure}
    \begin{subfigure}[b]{0.49\textwidth}
        \centering
        \includegraphics[width=\linewidth]{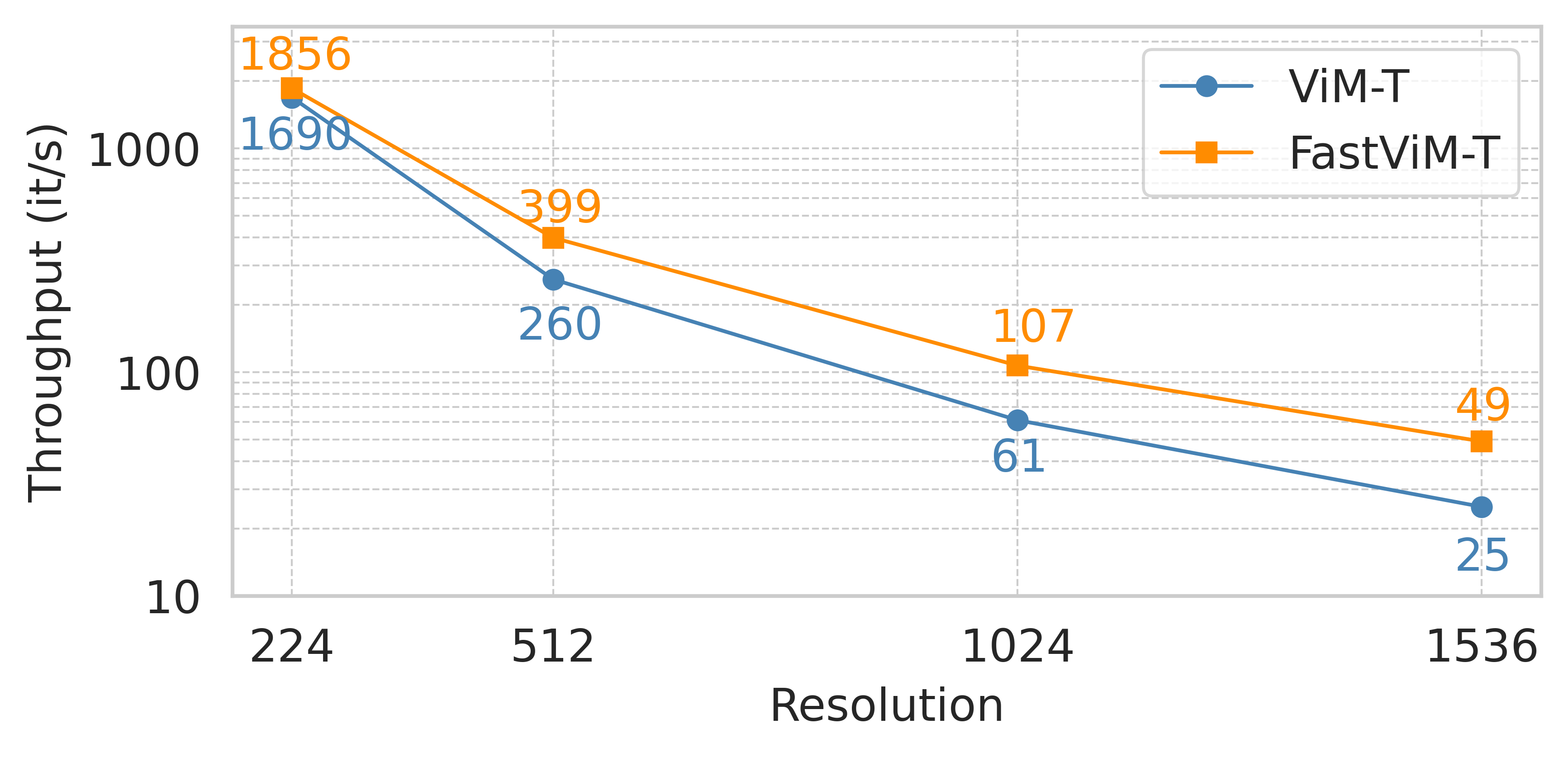}
        \caption{On A100 GPU}
    \end{subfigure}
    \caption{Comparison of Inference Throughput (it/s) for FastVim, Vim, and ViT across different resolutions. Tested with batch size of 128, with autocast as False, and with LayerNorm post-SSM in Vim and FastVim.}
    \label{fig:throughput_a100}
\end{figure*}

\begin{figure*}[!h]
    \centering
    \begin{subfigure}[b]{0.49\textwidth}
        \centering
        \includegraphics[width=\linewidth]{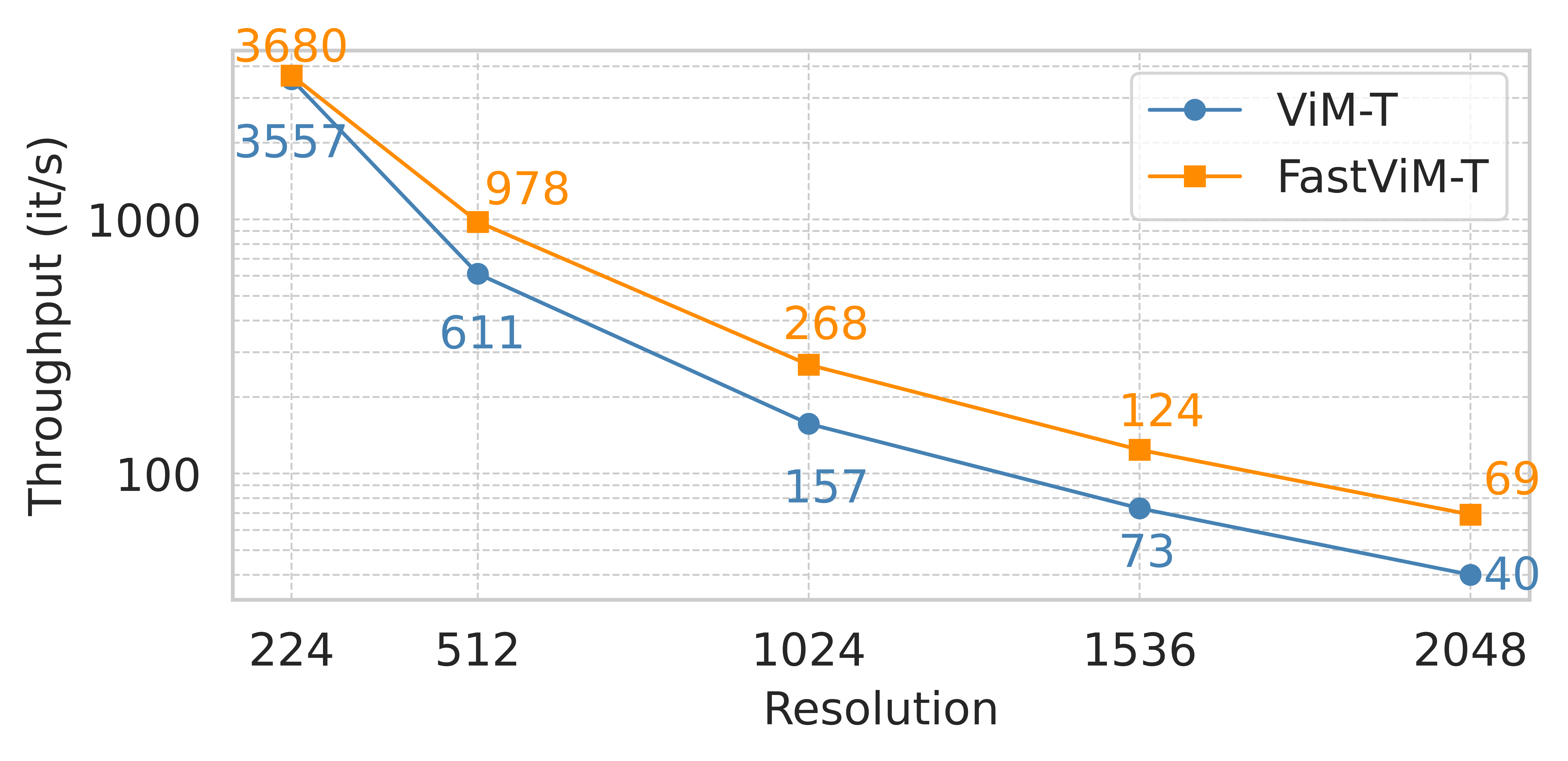}
        \caption{with LayerNorm post-SSM (default)}
        \label{fig:throughput_tiny}
    \end{subfigure}
    \begin{subfigure}[b]{0.49\textwidth}
        \centering
        \includegraphics[width=\linewidth]{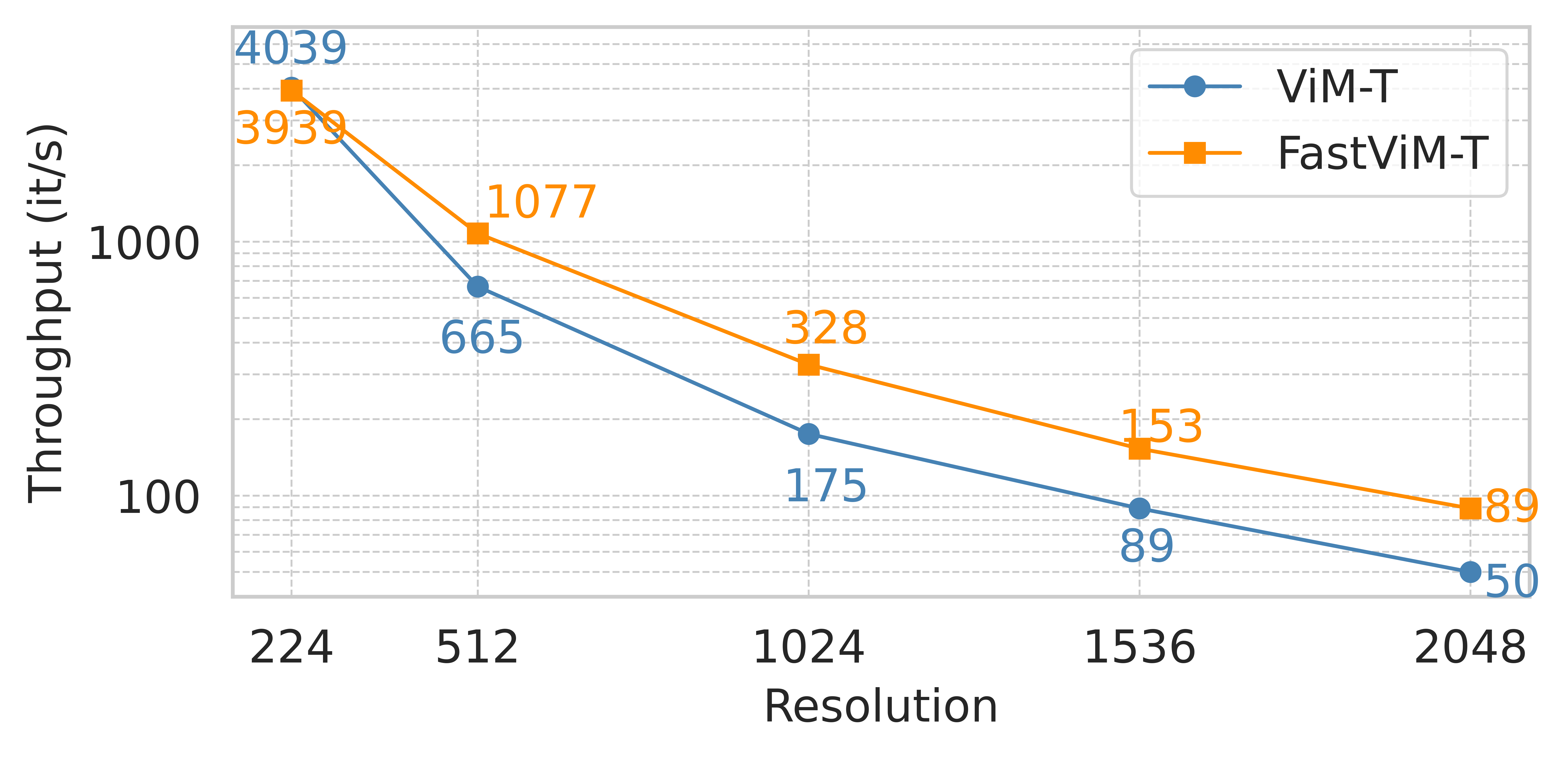}
        \caption{without LayerNorm post-SSM}
        \label{fig:throughput_base}
    \end{subfigure}
    \caption{Comparison of Inference Throughput (it/s) for FastVim, Vim, and ViT across different resolutions. Tested on H100 GPU with batch size of 128, and with autocast as False.}
    \label{fig:throughput_nolayernorm}
\end{figure*}

\begin{figure*}[!h]
    \centering
    \begin{subfigure}[b]{0.32\textwidth}
        \centering
        \includegraphics[width=\linewidth]{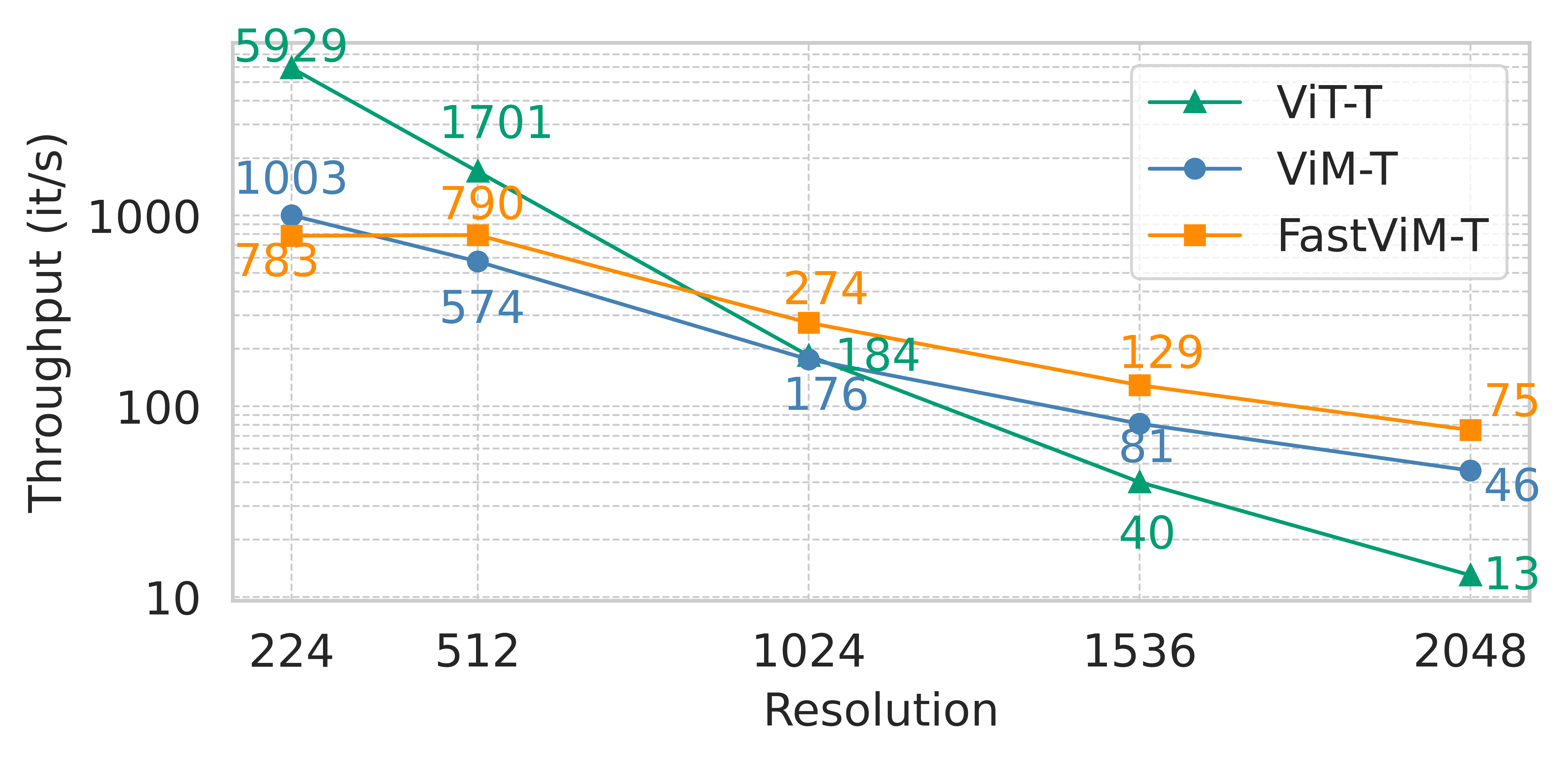}
        \caption{Tiny}
    \end{subfigure}
    \begin{subfigure}[b]{0.32\textwidth}
        \centering
        \includegraphics[width=\linewidth]{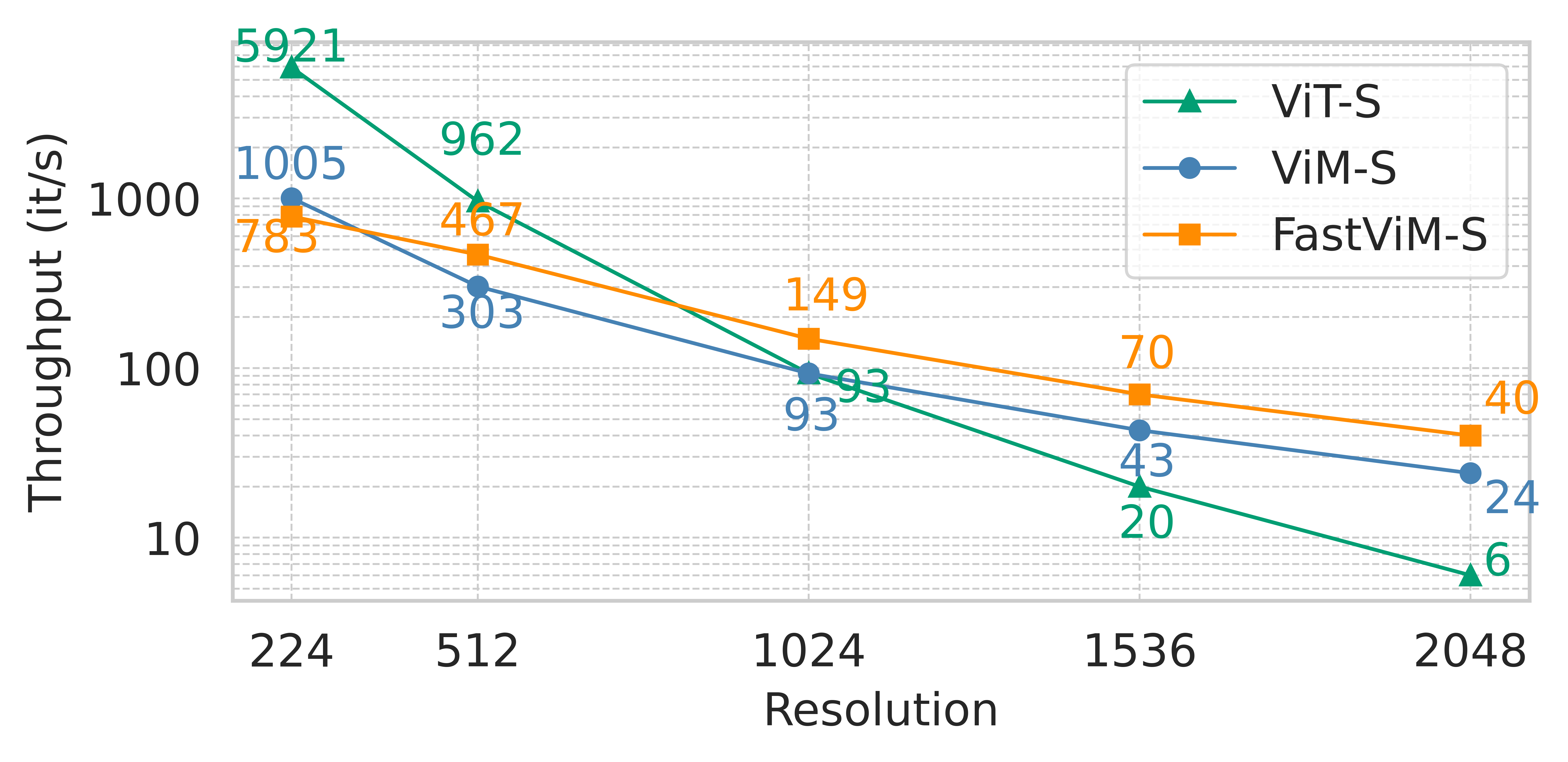}
        \caption{Small}
    \end{subfigure}
    \begin{subfigure}[b]{0.32\textwidth}
        \centering
        \includegraphics[width=\linewidth]{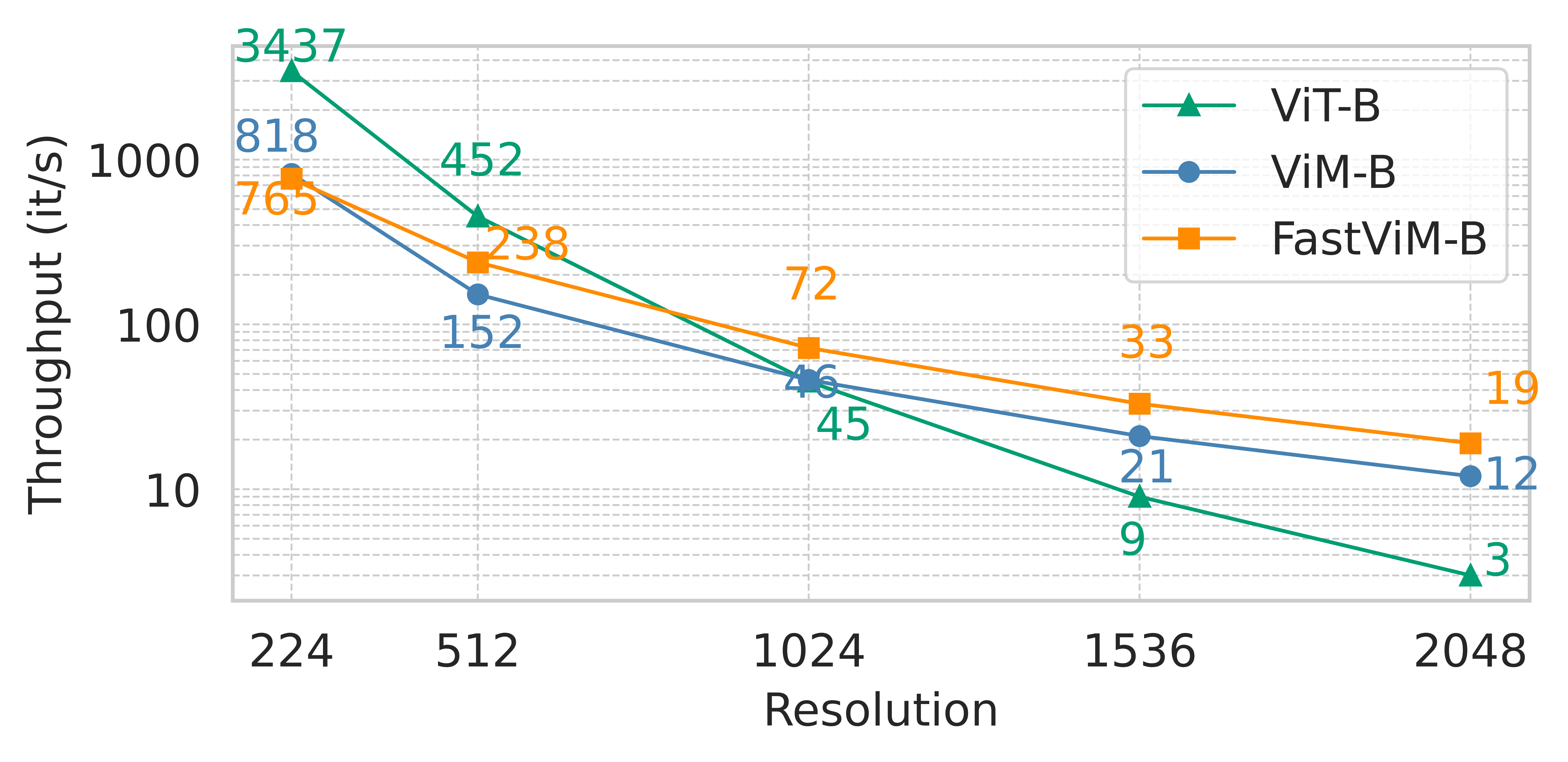}
        \caption{Base}
    \end{subfigure}
    \caption{Comparison of Inference Throughput (it/s) for FastVim, Vim, and ViT across different resolutions. Tested on H100 GPU with \textbf{batch size of 16}, autocast as False, and LayerNorm post-SSM in Vim and FastVim.}
    \label{fig:throughput_tiny_small_base}
\end{figure*}

\begin{enumerate}
    \item \textbf{Effect of Autocast.} In Fig.~\ref{fig:throughput_autocast_on}, we compare the throughput of ViT-T, Vim-T, and our FastVim-T across different resolutions, both with and without the autocast functionality for Vim and FastVim, since a few parameters need to be in floating point (fp) 32 in Mamba. In contrast, for ViT, we used fp16 directly. As illustrated, the Mamba-based methods showed little improvement in throughput with autocast enabled. However, ViT's throughput increased by approximately seven times with fp16 because FlashAttention-2~\cite{dao2023flashattention} is enabled at fp16, unlike in fp32. We would like to note that the goal of this study is to improve throughput in the Vision Mamba domain, which we have been able to achieve drastically. Further hardware-aware optimizations for our redundant repeat operation (see Sec.~\ref{kernel_details}) could provide additional improvements and potentially allow competition with ViTs in throughput speed at higher resolution, even with FlashAttention-2 enabled.
\\

    \item \textbf{Throughput on A100.} In Fig.~\ref{fig:throughput_a100}, we compare the throughput of Vim and FastVim on both A100 and H100 GPUs. As shown, at a resolution of 1536, FastVim provides almost a 100\% improvement on the A100 compared to a 70\% speedup on the H100 over Vim. The likely reason for this discrepancy is that our repeat operation, illustrated in Fig.~\ref{fig:model}, is computationally expensive and does not benefit significantly from the transition from A100 to H100. In contrast, other matrix operations become faster, resulting in a more pronounced improvement on the A100 GPU. 
\\

    \item \textbf{Effect of LayerNorm post-SSM.} In Fig.~\ref{fig:throughput_nolayernorm}, we illustrate the effect of using LayerNorm post-SSM on throughput for both Vim and FastVim. It is evident that adding LayerNorm results in slower throughput but is essential for maintaining stability, as shown in Fig.~\ref{fig:stability}. Unlike BatchNorm, LayerNorm requires computation even during inference, leading to a decrease in speed. However, previous works such as \textit{High-Performance Large-Scale Image Recognition Without Normalization}~\cite{brock2021high} and \textit{Vision Transformers Inference Acceleration Based on Adaptive Layer Normalization}~\cite{keddous2024vision} can be integrated to enhance FastVim's inference speed with the default setting of added LayerNorm post-SSM, without compromising stability.
\\

    \item \textbf{Throughput across model sizes.} In Fig.~\ref{fig:throughput_tiny_small_base}, we display the throughput of Vim and FastVim across Tiny, Small, and Base-sized models with a batch size of 16. Across all model sizes, our method consistently provides a speedup in throughput compared to the Vim baseline.
\\

    \item \textbf{Throughput on per-channel modeling tasks.} In Table~\ref{tab:throughput_channelvim}, we demonstrate the throughput improvement in FastChannelVim compared to ChannelVim. With a longer token sequence (patch size 8), FastChannelVim delivers a speedup of $62.3\%$ over ChannelVim without any drop in accuracy (see Table~\ref{tab:jumpcp_results}).
\\

\begin{table}[!h]
    \caption{Comparison of inference throughput analysis between ChannelVim and FastChannelVim across patch sizes 16 and 8. Autocast is set to false, and LayerNorm is added post-SSM. Each cell image has a resolution of 224 $\times$ 224 $\times$ 8, and the batch size is set to 8.}
    \begin{center}
    \resizebox{1\columnwidth}{!}{
    \begin{tabular}{lcc}
        \toprule
        Method & Token-grid & Throughput (it/s) \\
        \midrule 
         ChannelVim-S/16 & $14^2 \times$ 8  & 234 \\
         FastChannelVim-S/16 & $14^2 \times$ 8 & 318 \\

        \midrule
         ChannelVim-S/8 & $28^2 \times$ 8  & 61 \\
         FastChannelVim-S/8 & $28^2 \times$ 8 & 99 \\

            \bottomrule    
    \end{tabular}
}
\end{center}
    \label{tab:throughput_channelvim}
\end{table}

\begin{table*}[!h]
    \caption{Dissecting SSM processing time (in milliseconds) at inference for only the Forward and Backward SSM layer in one block of Vim-T versus FastVim-T. For Vim, SSM includes parameter projection + SSM scan (with skip connection in CUDA kernel); for FastVim, pool + parameter projection + SSM scan + repeat + skip connection. Note that since skip connection is added in CUDA kernel for SSM scan for Vim, we don't report the time for skip-conn. separately for Vim.}
    \begin{center}
    \resizebox{2\columnwidth}{!}{
    \begin{tabular}{c|cc|cc|cc|cc}
        \toprule
        Operations & Vim (224) & FastVim (224) & Vim (512) & FastVim (512) & Vim (1024) & FastVim (1024) & Vim (2048) & FastVim (2048) \\
    \midrule
        SSM scan  & 0.79 &  0.44 & 3.20 & 0.41   & 14.52 & 0.42   & 58.20 & 0.30 \\
         Parameter proj.  & 0.17 & 0.07 & 0.44 & 0.08 & 1.80 & 0.11 & 6.90 & 0.20 \\
         Pool  & - & 0.10 & - & 0.80 & - & 1.70 & - & 3.46 \\
         Repeat  & - & 0.06 & - & 0.26 & - & 1.00 & - & 3.90 \\
         Skip conn.  & - & 0.17  & - & 0.78  & - & 3.10  & - & 12.2 \\
\midrule
Total & 0.96 & 0.84 & 3.64 & 2.33 & 16.32 & 6.33 & 65.10 & 20.06  \\
            \bottomrule    
    \end{tabular}
}
\end{center}
    \label{tab:dissecting_SSM_time}
\end{table*}

\item \textbf{Dissecting SSM processing time.} Here, we calculate the processing time for Forward SSM + Backward SSM in only one block (see Fig.~\ref{fig:model}) for Vim-T versus FastVim-T. The SSM time include the parameter projection (\(\mathbf{B}, \mathbf{C}, \mathbf{\Delta}\)) for selective scan, the SSM \textbf{scan} time, and the skip connection (\(\mathbf{D} x_t\)). Note that since the Mamba implementation enables the skip connection inside the CUDA kernel for faster processing, for Vim, we put the skip connection inside the kernel. However, for FastVim, we can't input the skip connection matrix (\(\mathbf{D}\)) to the kernel since we need to first perform the repeat operation and then add it with the skip connection, which takes place outside the CUDA kernel in FastVim. This results in significant overhead for FastVim, but since our SSM scan and parameter projection is lot more computationally cheaper due to compressed input after pool, FastVim still gives significant speedup over Vim for the Forward SSM + Backward SSM in a block. 

\begin{figure}[!h]
\centering
    \includegraphics[width=1\linewidth]{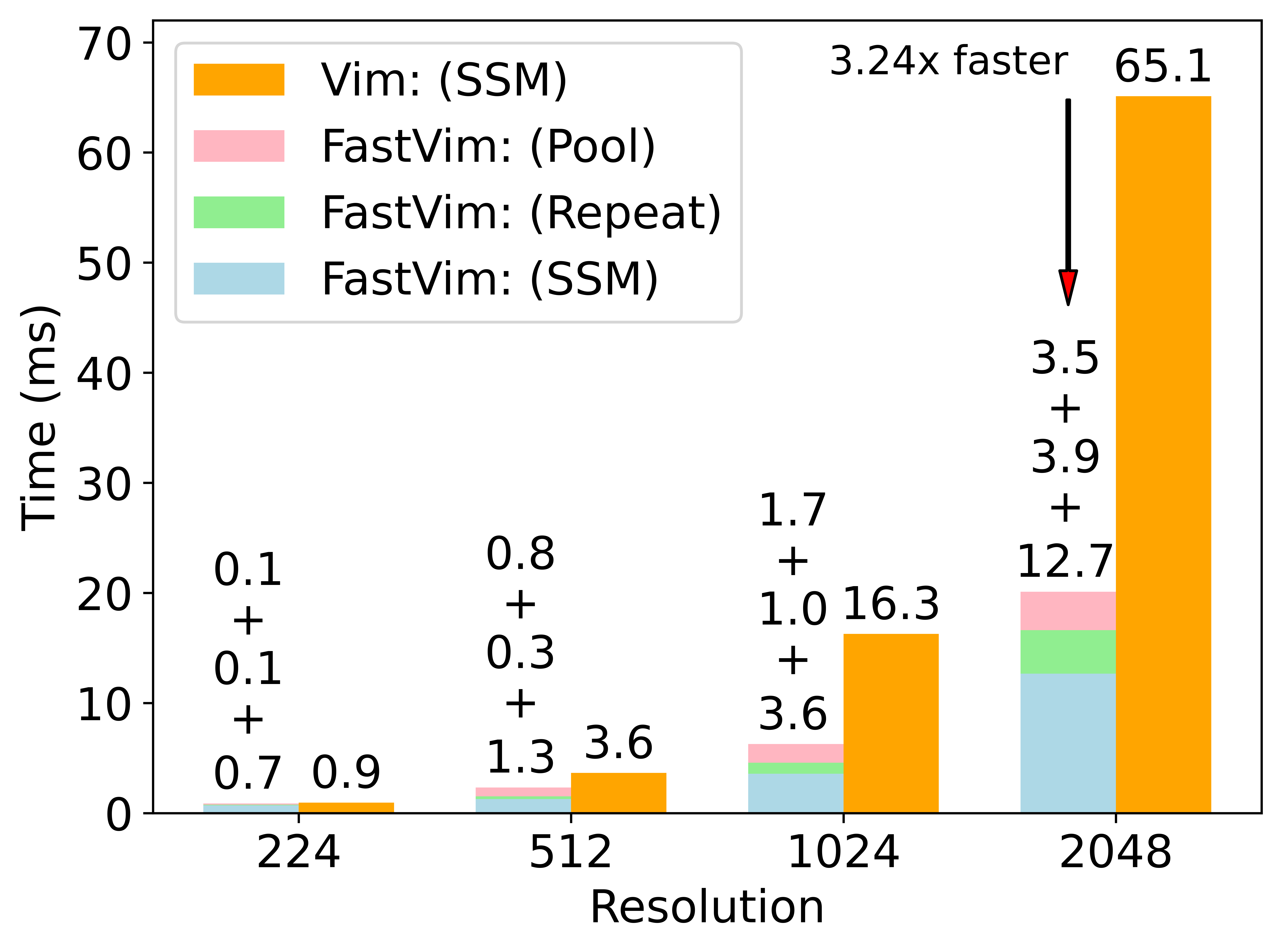} 
    \caption{Comparison of inference processing time for only the Forward and Backward SSM layer in one block of Vim-T versus FastVim-T. For FastVim, we calculate Pool + Repeat + SSM time. The annotations indicate SSM time for Vim; for FastVim, the upper value indicates the time for Pool, the middle value indicates the time for Repeat, and the lower value indicates the time for SSM.}
\label{fig:fastvim_vs_vim_ssm_plot}
\end{figure}

\end{enumerate}

In Fig.~\ref{fig:fastvim_vs_vim_ssm_plot}, for Vim-T, we calculate the time for SSM, whereas for FastVim-T, since we perform pooling and repeat operations, we measure the time for pooling and SSM and repeat. We can observe that the time taken in Vim scales quadratically with increasing resolution (approximately more than 4$\times$ increase in time for a 2$\times$ resolution increase), whereas ours scales sub-quadratically (approximately less than 3$\times$ increase in time for a 2$\times$ resolution increase). Thus at higher resolution (2048), FastVim (Pool + SSM + Repeat) is 3.24$\times$ faster than Vim (SSM). It can also be observed that as resolution increases, the repeat operation becomes increasingly expensive, taking almost 25\% of the time for FastVim (Pool + SSM + Repeat) compared to only about 8\% at resolution 224. Preliminary optimizations for this \textbf{redundant} repeat operation are discussed in Sec.~\ref{kernel_details}. Note that the time is recorded in milliseconds, and is for 1 batch with a batch size of 128. We use enough warmup runs to make sure the reported times are correct in practice. Residual (not the skip connection one) is omitted in these calculations for Vim and FastVim, and transpose operation is omitted as well in FastVim.
\\

In Table~\ref{tab:dissecting_SSM_time}, we mention the time taken by each component in more details for Forward SSM + Backward SSM for Vim and FastVim. For Vim, we report the parameter projection time and the SSM scan time. Note that since the Mamba kernel enables the skip connection in its CUDA kernel, for Vim, we do not separately report the skip connection time as it already becomes negligible in the Mamba kernel implementation. However, we observe that when the skip connection is not included inside the kernel, it takes significantly more time, similar to the skip connection time values mentioned for FastVim (in Table~\ref{tab:dissecting_SSM_time}). For FastVim, we measure the time for pooling, parameter projection, SSM scan, repeat operation, and skip connection, since it can't be added in the CUDA kernel due to the required repeat operation beforehand. It can be observed that even with a large image size of \(2048 \times 2048\), FastVim's SSM scan time and parameter projection time are lower than Vim's SSM scan time and parameter projection time at a much smaller 224 resolution. This is because, following tokenization and pooling, we have just \textbf{128} tokens for a 2048 resolution image, whereas Vim has 196 tokens for a 224 resolution image during the SSM scan and parameter projection. We would like to note that at higher resolutions, for FastVim, the pooling, repeat, and skip-connection operations take the majority of the time, whereas the SSM scan and parameter projection take significantly less time. These operations can be fused in CUDA kernel in future studies to achieve even more speedup.

\section{Semantic Segmentation implementation details}
\label{additional_semantic_implementation_section}

In line with Vim~\cite{vim} and LocalVim~\cite{huang2024localmamba}, we used a batch size of 16 and an input size of 512x512. We employed the AdamW optimizer with a weight decay of 0.01. A Poly learning rate schedule was used, decaying over 160K iterations, with an initial learning rate of \(6 \times 10^{-5}\). For Tiny and Small models, we used drop path rate of 0.05, and for Base, we used 0.4. For evaluation, we used sliding window prediction with crop size of 512 and stride of 341. We utilized the code provided by LocalVim~\cite{huang2024localmamba}, which is based on MMSegmentation~\cite{contributors2020openmmlab}.

\section{Object Detection and Instance Segmentation implementation details}
\label{additional_objectdet_implementation_section}

Following the code from LocalVim~\cite{huang2024localmamba}, we utilize the neck architecture from ViTDet and train Cascade Mask R-CNN as the detector. We employed the AdamW optimizer with a weight decay of 0.05, with a total batch size of 64. Initial learning rate is set to \(1 \times 10^{-4}\) and incorporates linear decay in the learning rate. We used drop path rate of 0.1 for Tiny and Small sized models, and 0.4 for Base sized model.

\section{Kernel details}
\label{kernel_details}

In FastVim (refer to Fig.~\ref{fig:model}), we apply mean pooling to the tokens before performing the SSM scan. Consequently, this operation must be repeated before integrating with the skip connection (\(\mathbf{D}\) in Eq.~\ref{eq:dtm}). When implementing this in PyTorch, we utilize the \texttt{repeat\_interleave} function to duplicate the output of the SSM scan prior to adding it with \(\mathbf{D} x_t\). However this operation becomes computationally expensive and redundant as demonstrated in Table~\ref{tab:dissecting_SSM_time}. Instead, we preliminarily explored modifying this repetition and moving the skip connection in the new CUDA kernel.

\begin{table}[!h]
    \caption{Comparison of inference throughput analysis with our kernel versus default Mamba kernel on a H100 gpu. Autocast is set to False, and LayerNorm is added post-SSM, image resolution is 224, and batch size is set to 128.}
    \begin{center}
    \resizebox{0.8\columnwidth}{!}{
    \begin{tabular}{lc}
        \toprule
        FastVim-T & Throughput (it/s) \\
        \midrule 
         with Mamba kernel  & 3680 \\
         with our kernel & 4009 \\

            \bottomrule    
    \end{tabular}
}
\end{center}
    \label{tab:throughput_with_kernel}
\end{table}

Specifically, given a input flattened token sequence ($x_t$) with a length $L = h \times w$, the compressed output (pool across column) will have a length of $h$. Our objective is to have the $i^{\text{th}}$ element of this compressed output directly added to the token sequence spanning \(i \cdot w\) to \((i+1) \cdot w\) within the skip connection \(\mathbf{D} x_t\). This technique can reduce the time spent on redundant repetition in a naive PyTorch implementation, translating to practical speedup. In Table~\ref{tab:throughput_with_kernel}, we demonstrate the increased throughput for FastVim at a resolution of 224 with our kernel compared to default Mamba kernel that we used in this study. With the current optimizations, we observe an improvement in speed for 224-sized images; however, there is a decrease in speed at higher resolutions. This indicates the need for further refinement to optimize our kernel. Therefore, we also plan to release our kernel implementation to the open-source community so that others can build upon it.

\section{Model configurations}
\label{model_sizes}

\begin{table}[!h]
    \caption{Model configurations for FastVim}
    \begin{center}
    \resizebox{0.6\columnwidth}{!}{
    \begin{tabular}{c|c|c}
        \toprule
        Model & Layers & Embedding dim. \\
    \midrule
         Tiny  & 24 & 192 \\
         Small  & 24 & 384 \\
         Base  & 24 & 768 \\
         Large  & 48 & 1024 \\
         Huge  & 64 & 1280 \\

            \bottomrule    
    \end{tabular}
}
\end{center}
    \label{tab:model_sizes}
\end{table}

\newpage

\end{document}